\documentclass[11pt,logo,onecolumn,copyright]{nvidiatechreport}

\usepackage[utf8]{inputenc}
\usepackage[T1]{fontenc}
\usepackage{amsmath,amssymb}
\usepackage{booktabs}
\usepackage{algorithm}
\usepackage{algpseudocode}
\usepackage{graphicx}
\usepackage{xcolor}
\usepackage{hyperref}
\usepackage{url}
\usepackage{upquote}
\usepackage{cleveref}
\usepackage{enumitem}
\usepackage{placeins}

\setlist{topsep=3pt,itemsep=1pt,parsep=1pt}
\newcommand{\method}{Decisive-Action BPB}

\newcommand{\passk}{\ensuremath{\mathrm{pass@}K}}
\newcommand{\passone}{\ensuremath{\mathrm{pass@}1}}
\newcommand{\mcqmethod}{Patch MCQ}

\usepackage[round]{natbib}

\title{Before They Can Solve:\\ Predicting Post-Training Coding-Agent\\ Performance from Base Models}

\author{
\mbox{Tan Yu$^{1}$}\quad
\mbox{Alexander Bukharin$^{1}$}\quad
\mbox{Khushi Bhardwaj$^{1}$}\quad
\mbox{Jennifer Williams$^{1}$}\quad
\mbox{Zirui Liu$^{1,2}$}\quad
\mbox{Jonathan Lingjie Li$^{1}$}\quad
\mbox{Soumye Singhal$^{1}$}\quad
\mbox{Joseph Jennings$^{1}$}\quad
\mbox{Sanjeev Satheesh$^{1}$}\quad
\mbox{Yash Jain$^{1}$}\quad
\mbox{Ashish Vaswani$^{1}$}\quad
\mbox{Venkat Krishna Srinivasan$^{1}$}\quad
\mbox{Matthew Papakipos$^{1}$}\quad
\mbox{Hyunwoo Kim$^{1}$}\quad
\mbox{Jian Zhang$^{1}$}\quad
\mbox{Oleksii Kuchaiev$^{1}$}\quad
\mbox{Markus Kliegl$^{1}$}\quad
\mbox{Mostofa Patwary$^{1}$}\quad
\mbox{Mohammad Shoeybi$^{1}$}\quad
\mbox{Bryan Catanzaro$^{1}$}\quad
\mbox{Jonathan Cohen$^{1}$}\quad
\mbox{Jiantao Jiao$^{1,3,\dagger}$}
\par\vspace{4pt}
{\footnotesize\textnormal{%
\mbox{$^{1}$NVIDIA}\quad
\mbox{$^{2}$University of Minnesota -- Twin Cities}\quad
\mbox{$^{3}$University of California, Berkeley}\\
\mbox{$^{\dagger}$Corresponding Author: \texttt{jiantaoj@nvidia.com}}
}}
}


\begin{document}
\begin{abstract}

How can we predict which base checkpoint is worth an expensive round of agentic post-training?
End-to-end \passk{} tests whether successful behavior already appears in a base model's distribution, but it is a poor fit for agentic coding: many base checkpoints cannot reliably produce the well-formed tool invocation required to complete a task end-to-end.
Single-shot or short-horizon tasks avoid these tool-calling failures by collapsing a multi-step interaction into a fixed prompt and a single patch, but they sidestep the core capability we care about: maintaining coherent state over many tool-using steps as the repository evolves.
To bridge this gap, we treat successful post-trained agent trajectories as a lookahead signal of base-model potential.
Replaying each trajectory and rerunning tests after every code-changing step identifies the \emph{decisive step}: the first step whose cumulative patch flips the repository from failing to passing, certifying that the recorded action solves the task given the prior context.
Motivated by a coverage principle for agentic traces, we build three screens at this step that do not require a base checkpoint to drive the harness from a cold start: (i) \method{} (bits per byte) measures the probability mass on the certified action, (ii) \mcqmethod{} tests the checkpoint's choice between that action and alternatives rejected by the same verifier, and (iii) prefix-conditioned \passk{} evaluates support for functionally-correct generations and credits any continuation that the tests accept.
Across ten pairs of public base and post-trained models, all three screens rank the cohort in close agreement with post-trained SWE-bench Verified \passone{}.
As our methods need only a benchmark’s successful trajectories and its verifier, they can be applied to turn future agentic coding benchmarks into base-model evaluations.
\end{abstract}

\maketitle

\section{Introduction}
\label{sec:intro}
Choosing a base checkpoint to post-train is a high-stakes decision made before its agentic capabilities can be fully realized. Revealing these capabilities requires substantial investment in compute, engineering, and evaluation, all of which comes only after the checkpoint has been chosen.

Existing evaluations fail to address this problem for opposite reasons. End-to-end agentic coding benchmarks such as SWE-bench Verified~\citep{jimenez2024swebench,openai2024verified} are too demanding for base checkpoints: \passk{} remains near zero because untuned base models fail to operate tool-use harnesses (Appendix~\ref{app:e2e}). Non-agentic coding benchmarks exercise isolated coding capabilities, but do not test sustained interaction with an evolving repository, and their correlation with post-trained agentic performance is highly inconsistent. Across ten pairs of open-source base models and their eventual post-trained checkpoints, Spearman $\rho$ with post-trained SWE-Bench Verified \passone{} ranges from $0.830$ on RepoBench XFirst to $-0.394$ on HumanEval (Appendix~\ref{app:bounded-baselines}). In short, current evaluations are either \emph{too agentic to run} or \emph{not agentic enough to predict downstream performance}.

In this paper, we address the question at the heart of this problem:
\textit{how can we evaluate agentic coding potential without requiring the base checkpoint to already behave like an agent?}
To answer this question, we build on the coverage principle: a promising base checkpoint should retain probability mass on behaviors that post-training may later select, rather than merely maximizing average correctness or conforming to today's agent protocol~\citep{chen2025coverage,yue2025does}. Agentic trajectories from frontier models on existing agentic coding benchmarks are a natural source for such desirable behaviors; an important follow-up is therefore \emph{where} in an agentic trajectory to probe that probability mass. We let the verifier of the original agentic coding task  decide.

\begin{figure}[t]
  \centering
  \includegraphics[width=0.98\textwidth]{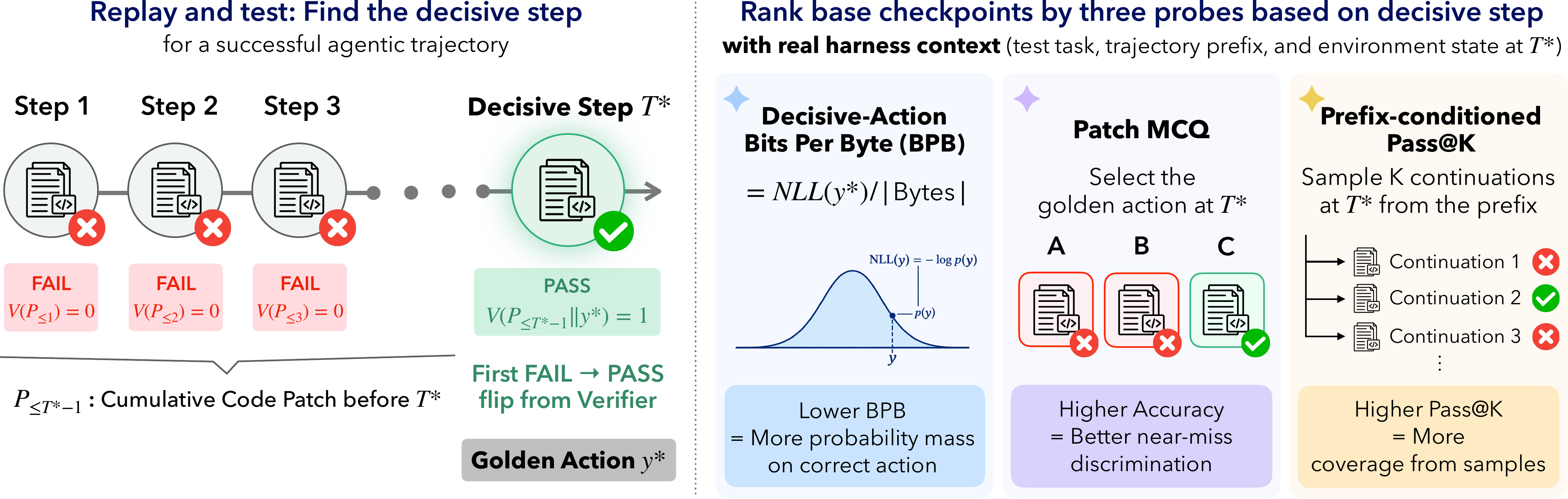}
  \caption{\textbf{A unified agentic trajectory-derived probing framework.} By replaying a coding-agent trajectory on a coding task against the task's own verifier $V$, we  locate the \emph{decisive step} $T^\star$, the earliest step whose cumulative patch resolves the task. 
  At this decisive step, (i) \method{} measures the byte-normalized likelihood which a base model assigns to the golden action; (ii) \mcqmethod{} evaluates base model's capability of differentiating the golden action from non-resolving alternatives; (iii) Prefix-conditioned \passk{} measures a base model's empirical success rate under sampling: it draws $K$ continuations from $T^\star$ and lets the task's own verifier $V$ judge them.}
  \label{fig:method-overview}
\end{figure}
For each step in a frontier post-trained LLM's successful trajectory for an agentic coding task, we reconstruct the cumulative environment state and run the task's tests at that state. This identifies the \emph{decisive step} $T^\star$, i.e., the first step at which the verifier accepts the solution, as visualized in Figure \ref{fig:method-overview}. 
The action taken at that step is our \emph{golden action}: conditioned on the preceding trajectory, executing it resolves the task, while the state immediately before it does not. Unlike the benchmark's \emph{gold patch} (i.e., the human-written reference fix), this is an action produced by a real coding agent and certified by the same verifier that defines task success.

At $T^{\star}$, we construct three complementary probes as visualized in Figure \ref{fig:method-overview}. \textbf{\method{}} asks an absolute question: how much probability mass does a checkpoint assign to the golden action? \textbf{\mcqmethod{}} asks a comparative question: can a checkpoint distinguish that action from plausible alternatives rejected by the verifier? \textbf{Prefix-conditioned \passk{}} asks a generative one: given the same trajectory prefix, can a checkpoint produce a continuation that passes the verifier? The three probes each answer a distinct, interpretable question about model behavior while holding the same agentic context and success criterion fixed. We validate them both by their aggregate correlation with post-trained agentic performance and by whether they correctly order individual checkpoint pairs by that same performance.

\noindent Our contributions are three-fold: 
\begin{itemize}[leftmargin=*,topsep=4pt,itemsep=2pt]
  \item \textbf{A framework for converting agentic coding benchmark into base-checkpoint evaluations.} We start from successful trajectories on an agentic benchmark, replay them, and run the task's tests after every code-changing step. The first step where tests pass is the decisive step: the action taken at that step becomes the golden action, and alternative edits attempted at the same point but rejected by the verifier become distractors. Each task yields (i) trajectory context until the decisive step (ii) a golden action (iii) distractors and (iv) an interactive, restorable environment state with an executable task verifier. The construction is a generic recipe rather than a fixed test set: any current or future agentic coding benchmark can be converted into a base-model evaluation.

  \item \textbf{Two static decision-point probes.} \method{} scores the base model by the likelihood it assigns to the golden action, whereas \mcqmethod{} scores a checkpoint by its choice among the verifier-certified golden action and the rejected distractors.

  \item \textbf{A dynamic continuation probe.} Prefix-conditioned \passk{} samples $k$ continuations from the decisive step and verifies each with the task's own tests; any edit that passes is accepted.
\end{itemize}

Across ten base checkpoints, all three probes produce rankings in close agreement with the ranking by post-trained performance on SWE-bench Verified. \method{} reaches Spearman $\rho=0.964$ on held-out trajectories from DeepSWE~\citep{huang2026deepswe}, a separate agentic coding benchmark. \mcqmethod{} reaches $\rho=0.903$ and prefix-conditioned \passk{} at $K=16$ reaches $\rho=0.988$ on trajectories drawn from SWE-bench Verified itself. Together, these results support our method of probing base checkpoints at verifier-certified decision points in real agent trajectories.

\section{Related Work}
\label{sec:related}
\textbf{End-to-end agentic evaluation.} Long-horizon, end-to-end agentic evaluations place a model inside an interactive environment and score whether the resulting model--harness system completes a task. SWE-bench Verified~\citep{jimenez2024swebench,openai2024verified} provides human-validated repository issues with test-based grading. SWE-bench Multilingual~\citep{yang2025swesmith} extends that task format to nine languages across 41 repositories. SWE-bench Pro~\citep{deng2025swebenchpro} targets longer-horizon, enterprise-style tasks. Terminal-Bench~\citep{merrill2026terminalbench} broadens the scope from repository issues to command-line workflows. DeepSWE~\citep{huang2026deepswe} uses original, multilingual tasks with purpose-built verifiers. FrontierCode~\citep{lu2026frontiercode} moves the bar from correctness to \emph{mergeability}, grading maintainer-authored tasks against an ensemble of unit tests. 
Some base checkpoint evaluation work treats end-to-end \passk{} as a capability ceiling~\citep{yue2025does}. However, these benchmarks are poorly suited to meaningfully evaluating base checkpoints that are usually too weak at such agentic capabilities.

\textbf{Non-agentic code evaluation.} 
Traditional code benchmarks isolate capabilities within bounded input--output tasks. HumanEval~\citep{chen2021evaluating} and MBPP~\citep{austin2021program} test function synthesis. EvalPlus~\citep{liu2023evalplus} provides more rigorous criteria for assessing the functional correctness of LLM-synthesized code. APPS~\citep{hendrycks2021apps} and DS-1000~\citep{lai2022ds1000} scale to competition problems and realistic libraries. BigCodeBench~\citep{zhuo2025bigcodebench} adds complex instructions and library use.  LiveCodeBench~\citep{jain2025livecodebench} emphasizes contamination-resistant competitive programming with one-step repair. CRUXEval~\citep{gu2024cruxeval} tests execution reasoning. RepoBench~\citep{liu2024repobench} and CrossCodeEval~\citep{ding2023crosscodeeval} add repository-level, cross-file context but retain a completion-style target. These benchmarks can be applied to base checkpoints, but do not test agentic behavior. 

\textbf{Base-checkpoint proxies for post-training potential.} APTBench~\citep{qin2025aptbench} and RuDE~\citep{li2026predicting} use answer discrimination to estimate base checkpoints’ downstream potential. RuDE contrasts answers labeled by model-judged rubric compliance; our agentic coding setting provides executable verifiers. APTBench’s SWE MCQs use reference-derived and LLM-extracted, rewritten, or generated answer choices, including gold patches paired with execution-filtered distractors. Our independent audit of 1,682 APTBench MCQs identifies questions whose designated correct option is incorrect or not uniquely correct (Appendix \ref{app:aptbench-audit}); Appendix \ref{app:aptbench-repro} reproduces its reported correlations on our cohort. Such defects weaken the interpretation of MCQ accuracy as the ability to distinguish correct from incorrect answers, even when aggregate scores correlate with post-training performance. We therefore construct \mcqmethod{} at the decisive step: the golden action is verifier-certified, while distractors are alternative actions rejected by the same verifier. This grounds the score in a specific, interpretable criterion: whether a checkpoint can distinguish the golden action from rejected plausible alternatives.

\textbf{Base-model support and post-training potential.} Base checkpoints can contain preference signal that their generations hide: pretrained models rank an answer with its true instruction far above a mismatched one even when their own samples are poor~\citep{hewitt2024instruction}. \citet{chen2025coverage} define the ``coverage'' principle, where strong 
base models already assign high probability mass to high-quality responses. RLVR mainly sharpens solutions base models can already sample rather than creating new ones~\citep{yue2025does}. Pushing further, \citet{wu2025invisibleleash} find RLVR improves \passone{}, yet with larger sampling budgets it loses access to more correct solutions than it gains, compared with base checkpoints before RLVR. Together, these results suggest base-model mass bounds post-training gains---exactly the quantity our probes measure. 

\textbf{Likelihood-based evaluation.} Likelihood has been studied heavily by the LLM community. \citet{huang2024compression}  show bits per character on domain corpora is near-linear in downstream benchmark performance~. In fact, per-domain loss correlated across public checkpoints has been shown to be strong enough to select pretraining data~\citep{thrush2024improving}. It is also smoother than task accuracy, giving a more usable development signal than discontinuous task scores~\citep{heineman2025signal}. Raw likelihood is strongly affected by sequence length, confounding comparisons unless scores are length-normalized~\citep{holtzman2021surface}. Bits per byte (BPB) is one normalization that keeps checkpoints with different tokenizers comparable~\citep{magnusson2024paloma}. We adopt BPB when constructing our evaluation framework. 
\section{Trajectory-Derived Protocols}
\label{sec:method}
\label{sec:theory}
In this section, we define how we convert agentic coding benchmarks into base checkpoint evaluations. We define the decisive step in~\Cref{sec:decisive-discovery}  and then use it to construct three probes: Decisive-Action BPB (\Cref{sec:cbpb-def}), Patch MCQ (\Cref{sec:mcq-def}), and prefix-conditioned pass@$K$ (\Cref{sec:passk-def}).

\subsection{decisive-step discovery for agentic trajectories}
\label{sec:decisive-discovery}

Given a task from an agentic coding benchmark, \emph{e.g}, SWE-Bench, we collect a successful trajectory of $L$ steps from a frontier post-trained LLM using an agentic coding harness, \emph{e.g.,} mini-swe-agent. Let $C \subseteq \{1,\ldots,L\}$ be the indices for the set of steps involving code changes. Replaying through $T \in C$ yields a cumulative code patch $P_{\le T}$. We define the verifier-derived \emph{decisive step} as
\begin{equation}
\label{eq:desisive_turn}
    T^\star := \min\{T \in C : V(P_{\le T}) = 1\},
\end{equation}

where $V(\cdot) \in \{0,1\}$ denotes the verifier for the task.
As we collected a successful trajectory, it is guaranteed that a step satisfying the above condition exists. The decisive step identifies the evaluation boundary: the evaluation context contains only steps strictly before $T^\star$.

\begin{figure}[t]
  \centering
  \includegraphics[width=0.9\textwidth]{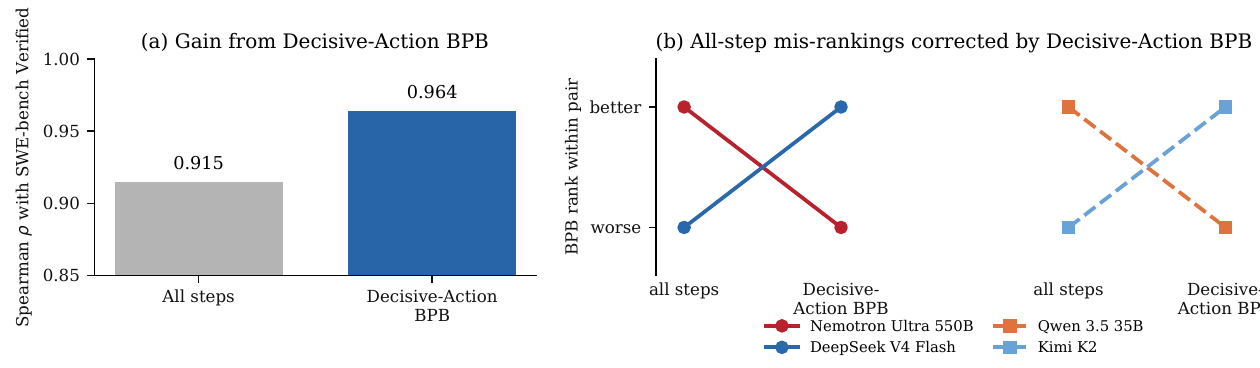}
  \vspace{-1mm}
  \caption{\textbf{(a)} Restricting BPB to the verifier-derived decisive step, instead of averaging over every step, raises its Spearman's rank correlation ($\rho$) with downstream agent capability, from 0.915 to 0.964 against SWE-bench Verified, across the same ten checkpoints. \textbf{(b)} At the checkpoint level, Decisive-Action BPB ranks DeepSeek V4 Flash above Nemotron Ultra and Kimi K2 above Qwen 3.5 35B, in agreement with their downstream performance, thus reversing the incorrect orderings produced by all-step BPB.}
  \label{fig:teaser}
\end{figure}

\textbf{The golden action.} Let $y^{\star}$ denote the LLM response in step $T^{\star}$ of the successful trajectory. Thus, by the definition of decisive step $T^{\star}$ in Eq.~\ref{eq:desisive_turn},
\begin{equation}
V\!\left(P_{\leq T^\star -1}\right) = 0, \qquad V\!\left(P_{\leq T^\star -1}  \Vert  y^{\star}\right) = 1,
\label{eq:golden}
\end{equation}
where $P_{\leq T^\star -1}  \Vert  y^{\star}$ denotes the cumulative patch obtained by applying the action $y^{\star}$ on  $P_{ \leq T^\star -1}$. We call $y^{\star}$ \emph{golden given its prefix}. Before $T^\star$, none of the recorded actions induces a fail-to-pass transition, so rewarding agreement with them mainly measures imitation of the trajectory generator (or incidental edits) rather than problem-solving ability. At $T^\star$, applying $y^{\star}$ causes the verifier to accept the cumulative patch for the first time. After $T^\star$, later actions may introduce changes unrelated to resolving the task. Therefore, only the decisive step yields a verifier-certified golden action, whereas the steps before and after it are not golden. \Cref{fig:teaser} shows that this choice is not merely principled: restricting BPB to the decisive step increases rank agreement with downstream agent capability, and reverses two checkpoint pairs that all-step BPB orders incorrectly.

\subsection{\method{}}
\label{sec:cbpb-def}
\method{} measures how much probability mass a checkpoint assigns to the golden action, given its prefix. Let $x^{\star}$ be the trajectory prefix before the decisive steps' golden action, $y^{\star}$. We use one fixed universal chat template and tokenize the concatenation $x^{\star}  \Vert  y^{\star}$ into $t$. We then compute $L(\theta)$, the golden action's negative log-likelihood in bits for the tokens fully contained in  $y^{\star}$, and normalize it by the action's byte length, $|B|$, to obtain BPB. This normalization makes scores comparable across checkpoints using different tokenizers~\citep{gao2020pile,magnusson2024paloma}. Let $S$ denote the set of token indices fully contained in $y^{\star}$ and $B$ denote the set of bytes belonging to $y^{\star}$; \method{} is then formulated as
\begin{equation}
L(\theta) := -\!\!\sum\nolimits_{j \in S}\!\log_2 p_\theta(t_{j} \mid t_{1:j-1}), \qquad \operatorname{BPB}(\theta) := L(\theta)/\lvert B \rvert.
\label{eq:bpb}
\end{equation}
A lower BPB is better, meaning the checkpoint places more probability mass on $y^{\star}$ .

\textbf{Chat Template Masking.} Base models may not be familiar with the simple chat template we use for BPB calculations. To reduce the impact of template choice on the results, we mask out formatting text added by the chat template when computing BPB (Appendix \ref{app:chat-template}).

\subsection{Patch MCQ}
\label{sec:mcq-def}
Decisive-Action BPB alone cannot establish whether an action resolves a task: two actions can receive similar scores even when a single-character difference causes one to pass verification and the other to fail. Our patch-based multiple-choice probe (\mcqmethod{}) addresses this limitation through a discriminative $k$-way choice, presenting the golden action alongside alternatives that fail verification when applied after the same trajectory prefix. The probe asks the checkpoint to select the action that resolves the task, rather than only measuring the likelihood of the golden action.

Conditioned on trajectory prefix $x^\star$, we sample a set of candidate actions from post-trained LLMs, denoted $\mathcal{Y}$. We label each candidate by applying it on the prefix’s cumulative patch $P_{\leq T^\star-1}$ and checking the verifier outcome. The distractor set, $\mathcal{Y}^-$, comprises the candidates that fail verification:

\begin{equation}
   \mathcal{Y}^{-} = \{y: y \in \mathcal{Y} \land V(P_{\le T^\star -1}  \Vert  y) = 0 \}.
\label{eq:mcq-options}
\end{equation}

Thus, every distractor is identified by its failure to pass the task's verifier rather than by a human or model-based rubric. While, the golden action $y^{\star}$ in the original  trajectory, is the correct option. 

We score all options jointly through answer-letter likelihood. The MCQ context $u$ contains the trajectory prefix and complete option set.  Let $\mathcal{L}_k := \{\ell_1,\ldots,\ell_k\}$ denote the answer-letter tokens assigned to the $k$ displayed options (for our four-option panels, $\mathcal{L}_4=\{\mathrm{A},\mathrm{B},\mathrm{C},\mathrm{D}\}$). We select the most likely next-token answer letter:
\begin{equation}
\widehat{\ell}
:= \arg\max_{\ell \in \mathcal{L}_k} p_\theta(\ell \mid u),
\label{eq:mcq-letter-likelihood}
\end{equation}
Thus, answer-letter likelihood compares all displayed actions within one shared context rather than scoring each action independently.
Let $\mathcal{Q} = \{q_i\}_{i=1}^N$ denote the set of constructed multiple-choice questions. For each question $q_i$, let  $\hat{\ell}_i$ denote the predicted option and $\bar{\ell}_i$ denote the golden option. The patch MCQ accuracy $r$ is computed by

\begin{equation}
 r =  \frac{1}{N} \sum_{i=1}^n \mathbf{1}(\hat{\ell}_i = \bar{\ell}_i),
\end{equation}
 where $\mathbf{1}(\cdot) \in \{0,1\}$ is  indicator function.  To reduce positional bias, we use cyclic option reordering~\citep{zheng2024large}. The answer probabilities are aggregated across reorderings before selecting the predicted answer, $\widehat{\ell}$, detailed implementation details are in Appendix~\ref{app:mcq-detail}. 

\subsection{Prefix-conditioned \texorpdfstring{\passk{}}{pass@k}}
\label{sec:passk-def}
\method{} and \mcqmethod{} both score the base checkpoint's distribution over actions that were previously generated, but neither tests whether the checkpoint can successfully generate resolving actions. However, on agentic coding tasks, end-to-end \passk{} for base models is typically zero even for very large $K$, since success often requires successive tool calls and agentic interactions with environments, which are not achievable by base models before post-training. \emph{Prefix-conditioned \passk{}} measures whether at least one of $K$ sampled continuations from the trajectory prefix $P_{i,\leq T_i^{\star}-1}$ successfully resolves the task.

 We therefore condition on real trajectories as context: for each successful trajectory $i$ from a frontier post-trained LLM, let $T_i^{\star}$ be its decisive step and let $P_{i,\leq T_i^{\star}-1}$ denote the cumulative patch as identified in \cref{sec:decisive-discovery}. 
 We let $x^{\star}_i$ be the trajectory prefix before the decisive step's golden action  and  $\hat y_{i,j}$ be a sampled continuation from a base checkpoint. \emph{Prefix-conditioned \passk{}} measures whether at least one of $K$ sampled continuations from the base checkpoint  resolves the task:

\begin{equation}
    \operatorname{\passk{}}(\theta) := \frac{1}{|\mathcal{I}|}\sum_{i \in \mathcal{I}} \mathbf{1}\!\left[\exists\, j \le K : V_i\!\left(P_{i,\leq T_i^{\star}-1} \,\Vert\, \hat y_{i,j}\right) = 1\right].
\end{equation}
Appendix~\ref{app:passk-detail} describes the rollout mechanics, including the masking convention at $T_i^\star$, the forward budget, how a base checkpoint is bridged onto the agent interface, and how the $k$ rollouts are verified.

\section{Experiments}
\label{sec:setup}
We evaluate all three trajectory-derived probes on the same ten base checkpoints. We  measure how strongly each proposed metric: \method{}, \mcqmethod{}, and prefix-conditioned \passk{}, correlates with models' relative agentic capability after post-training. We use \passone{} performance on SWE-bench Verified~\citep{jimenez2024swebench,openai2024verified}, SWE-bench Multilingual~\citep{yang2025swesmith}, and Terminal-Bench 2.1~\citep{merrill2026terminalbench} as proxies for post-training agentic capability.

\subsection{Methodology}
\label{sec:exp-method}
\FloatBarrier

The three probes are built on trajectories collected from GPT-5.6-Sol~\citep{openai2026gpt56sol}, Opus-5~\citep{claude_opus_5_5}, and Kimi-K3~\citep{team2026kimi} on SWE-Bench Verified, SWE-Pro, and DeepSWE, using mini-swe-agent~\citep{yang2024sweagent} as the harness. 
The distractors  in Patch MCQ are generated from a  set of LLMs including Opus-4.8~\citep{claude_opus_4_8}, Sonnet-5~\citep{claude_sonnet-5}, GPT-5.6-Terra~\citep{openai2026gpt56terra}, GPT-5.6-Luna~\citep{openai2026gpt56luna},  
GLM-5.2~\citep{glm5team2026glm5vibecodingagentic}, Gemini-3.7-Flash~\citep{google2026gemini37flash} and Gemini-3.6-Flash~\citep{google2026gemini36flash}.

We probe a total of ten base checkpoints: Nemotron-3-Nano~\citep{blakeman2025nemotron}, Nemotron-3-Super~\citep{chandiramani2026nemotron}, Nemotron-3-Ultra~\citep{blakeman2026nemotron}, Qwen-3.5-35B-A3B-Base~\citep{qwen3.5}, DeepSeek-V4-Flash-Base~\citep{deepseekai2026deepseekv4}, Kimi-K2-Base~\citep{team2025kimi}, GLM-4.5-Air-Base~\citep{zeng2025glm}, Hy3-Preview-Base~\citep{tencent2026hy3}, DeepSeek-V4-Pro-Base~\citep{deepseekai2026deepseekv4}, and Gemma-4-26B-A4B~\citep{team2026gemma}.

For each metric, we measure its correlation across the ten base checkpoints with the SWE-bench Verified and Terminal-Bench 2.1 performance of the corresponding post-trained checkpoints. We use publicly reported benchmark scores when available and run the evaluations ourselves when they are not. The full pipeline for computing each metric, from trajectory collection through verification, is described in Appendix~\ref{app:infra}. Further experimental details are provided in Appendix~\ref{app:setup-detail}.

We correlate each probe's per-checkpoint score $s$ (using $s=-\operatorname{BPB}$ for \method{}, so higher is better) with $d$, the downstream \passone{} of the checkpoint's post-trained family. Pearson $r=\operatorname{cov}(s,d)/(\sigma_s\sigma_d)$ measures linear agreement; Spearman $\rho$, Pearson $r$ on ranks, measures whether a probe orders checkpoints correctly, which is all checkpoint selection needs.

\begin{figure}[!ht]
  \centering
  \includegraphics[width=0.92\textwidth]{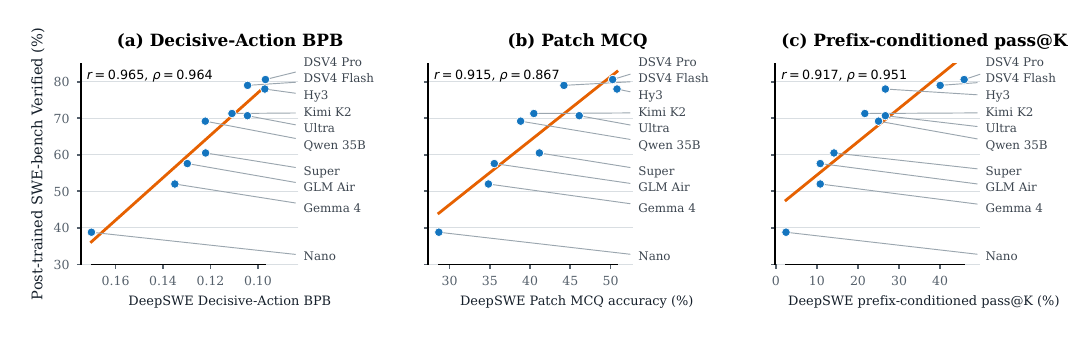}
  \caption{\textbf{All three DeepSWE-derived probes strongly track post-trained SWE-bench Verified performance.} The common DeepSWE source corpus makes the panels directly comparable across \method{} with tool-formatting tokens masked,  \mcqmethod{}, and prefix-conditioned \passk{} at $K=32$. The same ten checkpoints appear in every panel; annotations report Pearson $r$ and Spearman $\rho$. The BPB axis is reversed so rightward consistently means better.}
  \label{fig:deepswe-probes-swe-verified}

  \vspace{6pt}
  
  \centering
  \includegraphics[width=0.92\textwidth]{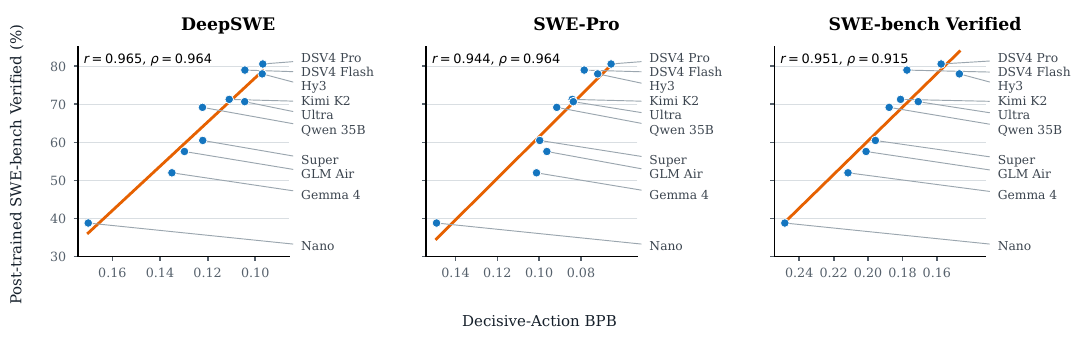}
  \caption{\textbf{Decisive-Action BPB strongly tracks post-trained SWE-bench Verified performance.} Each panel measures the same ten base checkpoints on a different trajectory corpus. Axes run from higher to lower BPB so rightward means better; annotations report Pearson $r$ and Spearman $\rho$ with $-\operatorname{BPB}$.} 
  \label{fig:decisive-bpb-scatter}

  \vspace{6pt}

  \centering
  \includegraphics[width=0.92\textwidth]{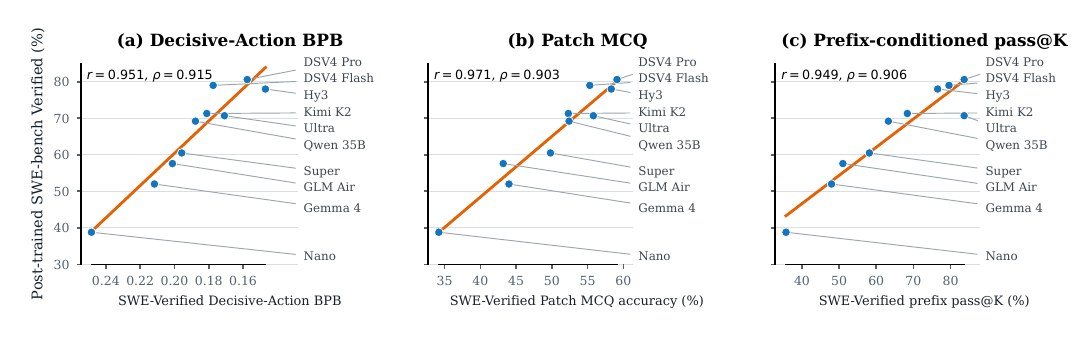}
  \caption{\textbf{All three probes also track post-trained SWE-bench Verified performance when constructed from SWE-bench Verified trajectories.} Panels use the same ten checkpoints and settings as \cref{fig:deepswe-probes-swe-verified}, but replace the DeepSWE source trajectories with SWE-bench Verified trajectories. Annotations report Pearson $r$ and Spearman $\rho$; the BPB axis is reversed so rightward consistently means better. Because the source and downstream target use the same benchmark, this is same-benchmark rather than cross-benchmark evidence.}
  \label{fig:swev-probes-swe-verified}

\end{figure}
\subsection{Correlations with SWE-Verified}
\label{sec:exp-swev-corr}

\Cref{fig:deepswe-probes-swe-verified} shows that each of our probes strongly correlates with downstream SWE-bench Verified performance. While we choose DeepSWE as the source task for probe construction in \Cref{fig:deepswe-probes-swe-verified}, these strong correlations hold across trajectories collected from different tasks, as shown in \Cref{fig:decisive-bpb-scatter}.

\Cref{fig:swev-probes-swe-verified} repeats the analysis with probes built from SWE-bench Verified trajectories instead. All three still track post-trained performance closely, at Spearman $\rho$ of 0.915 for \method{}, 0.903 for \mcqmethod{}, and 0.906 for prefix-conditioned \passk{}. Matching the source to the target benchmark does not, however, buy higher rank agreement. For \method{} and prefix-conditioned \passk{}, the held-out DeepSWE source agrees more closely ($\rho=0.964$ versus 0.915, and 0.951 versus 0.906), and only \mcqmethod{} gains from the match (0.903 versus 0.867).

\textbf{Bounded code benchmarks do not recover the post-trained ranking.} \Cref{tab:bounded-baselines-app} in \cref{app:bounded-baselines} correlates the same ten checkpoints' scores on bounded code benchmarks with the same downstream targets. Synthesis benchmarks carry almost no rank signal (MBPP $\rho=0.091$, HumanEval $\rho=-0.394$), and even the strongest baseline, RepoBench XFirst ($\rho=0.830$), trails all three DeepSWE-derived probes ($\rho$ of 0.867--0.964).

\FloatBarrier
\subsection{Correlations with SWE-Multilingual}
\label{sec:exp-swemult-tb-corr}
\FloatBarrier

\begin{figure}[!htbp]
  \centering
  \includegraphics[width=0.92\textwidth]{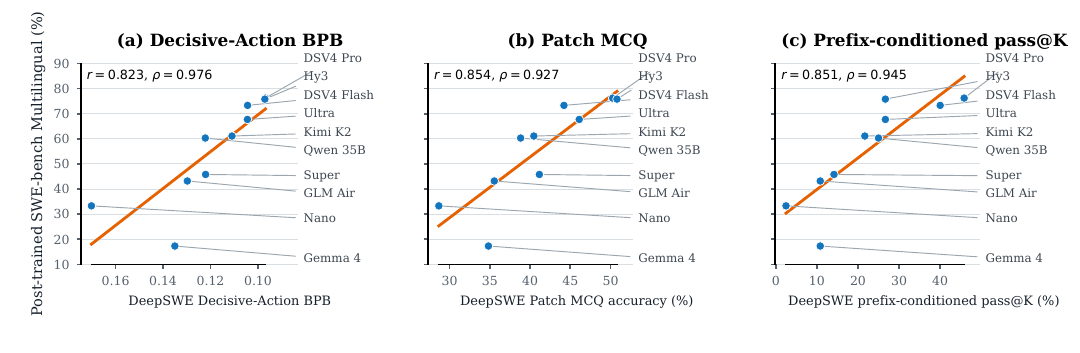}
  \caption{\textbf{The three DeepSWE-derived probes retain strong rank agreement with post-trained SWE-bench Multilingual performance.} Panels use the same ten checkpoints and the same probe settings as \cref{fig:deepswe-probes-swe-verified}; annotations report Pearson $r$ and Spearman $\rho$. The BPB axis is reversed so rightward consistently means better.}
  \label{fig:deepswe-probes-swe-multilingual}
\end{figure}

\Cref{fig:deepswe-probes-swe-multilingual} tests whether the same DeepSWE-derived probes predict post-trained performance beyond SWE-bench Verified. On SWE-bench Multilingual, rank agreement holds or improves. All three probes keep strong rank agreement with SWE-bench Multilingual, at $\rho=0.976$ for \method{}, 0.927 for \mcqmethod{}, and 0.945 for prefix-conditioned \passk{}; the first is the highest of the nine probe--target pairs in \cref{fig:deepswe-probes-swe-verified,fig:deepswe-probes-swe-multilingual,fig:deepswe-probes-terminal}, and \mcqmethod{} agrees more closely here than with SWE-bench Verified (0.867). One plausible contributor is that the DeepSWE source is itself multilingual. Pearson $r$ is noticeably lower, 0.82--0.85, and the gap between the two statistics has a single cause: Gemma 4 26B scores 17.3\% on SWE-bench Multilingual against 52.0\% on SWE-bench Verified. Correlations with Terminal-Bench 2.1 are reported in \cref{app:terminal-bench}.

\subsection{SFT Experiments}
\label{sec:exp-sft}
\FloatBarrier

\begin{table}[h!tbp]
  \centering
  \small
  \setlength{\tabcolsep}{4pt}
  \begin{tabular}{@{}lcccccc}
    \toprule
    & \multicolumn{1}{c}{Post-SFT} & \multicolumn{1}{c}{Public} & Decisive-Action & Patch & Prefix \\
    Base checkpoint & \multicolumn{1}{c}{SWE-Verified} & \multicolumn{1}{c}{SWE-Verified} & BPB & MCQ & pass@32 \\
    \midrule
    Nemotron-3-Nano-Base & 28.5 & 38.8 & 0.149 & 33.4 & 13.1 \\
    Nemotron-3-Super-Base & 53.1 & 60.5 & 0.100 & 34.4 & 34.6\\
    Qwen-3.5-35B-A3B-Base  & \textbf{61.2} & \textbf{69.2} & \textbf{0.092} & \textbf{36.2} & \textbf{46.9} \\
    \bottomrule
  \end{tabular}
  \caption{Under the same SFT procedure, post-SFT SWE-bench Verified follows the same ordering as the publicly reported scores, and the probes order the checkpoints consistently with it. Probe results are computed with SWE-Pro trajectories.}
  \label{tab:probes-sft}
\end{table}

\noindent \textbf{Under a controlled SFT recipe, the correlation trend continues.}
The downstream performance results in Sections \ref{sec:exp-swev-corr} and \ref{sec:exp-swemult-tb-corr} are computed using publicly released post-trained models, each produced by a different lab with their own post-training recipe and evaluation harness. 
To isolate the contribution of the base checkpoint, we fine-tune three base checkpoints, namely, Nemotron-3-Nano--Base, Nemotron-3-Super-Base, and Qwen-3.5-35B-A3B-Base, with an identical SFT recipe. 
Each model is trained on the same internal corpus for 1{,}000 steps, amounting to 16.8B training tokens. \Cref{tab:probes-sft} shows the resulting checkpoints' performance on SWE bench Verified, controlled for the same harness and evaluation hyperparameters. The post-SFT ordering agrees with the publicly reported one, and all three probes computed on SWE-Pro trajectories order the checkpoints the same way. With other trajectory sources the agreement is weaker for some probes (\Cref{tab:probes-sft-corr} in Appendix~\ref{app:sft-corr}), and with only three checkpoints we read this as the correlation trend continuing under a controlled recipe rather than as a precise estimate.

\subsection{Ablation Studies}
\label{sec:exp-ablations}

\begin{figure}[H]
  \centering
  \includegraphics[width=0.92\textwidth]{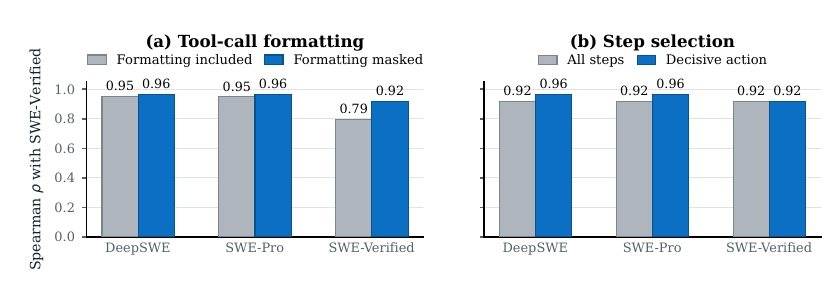}
  \caption{\textbf{Masking tool-call formatting and scoring only decisive actions improve the BPB screen.} Bars report Spearman correlation with post-trained SWE-bench Verified; all correlations use $-\operatorname{BPB}$ so higher is better.}
  \label{fig:bpb-ablation-summary}
\end{figure}

\textbf{Decisive step ablation.} 
In section \ref{sec:decisive-discovery}, we hypothesize that decisive steps are the best parts of the trajectory to evaluate. We further examine 
the importance of decisive steps in (\Cref{fig:bpb-ablation-summary}b), where we find decisive step BPB correlates much better than all turns.

\textbf{Format masking ablation.} We ablate whether chat-template tokens (specifically the tool-call formatting) contribute to Decisive-Action BPB, holding the trajectory, action, and semantic content fixed. Masking the tool-call formatting improves Spearman agreement on all our corpora (\Cref{fig:bpb-ablation-summary}a). The difference is especially large for SWE-verified traces. We hypothesize that this is due to the fact that SWE-verified often has early decisive actions. At these early steps the base model has not seen a lot of prior tool-call formatting examples, causing BPB spikes during evaluation. These findings support our methodology in Section \ref{sec:cbpb-def}.

\textbf{\mcqmethod{} depends on scoring all options jointly.} \Cref{tab:mcq-downstream-app} ablates how \mcqmethod{} selects an answer. Scoring each option independently and selecting the lowest BPB (argmin-BPB) shows only weak agreement with post-trained SWE-bench Verified ($\rho=0.665$ on DeepSWE questions; $\rho=0.578$ on SWE-Verified questions). Presenting all four options in a single context and reading the answer-letter likelihood under cyclic rotation increases agreement to $\rho=0.806$ and $\rho=0.842$ without reasoning; allowing reasoning before the answer yields smaller additional gains (to 0.867 and 0.903). Thus, joint scoring explains most of the agreement on both question sets, while reasoning provides a modest refinement.
Detailed results are in Appendix \ref{app:mcq-detail}.

\begin{table}[h]
  \centering
  \footnotesize
  \begin{minipage}[t]{0.48\textwidth}
    \begin{tabular*}{\linewidth}{@{\extracolsep{\fill}}lrr@{}}
      \toprule
      Scoring method & DeepSWE & SWE-Verified \\
      \midrule
      Argmin-BPB & 0.665 & 0.578 \\
      Rotated, no think & 0.806 & 0.842 \\
      Rotated, think & 0.867 & 0.903 \\
      \bottomrule
    \end{tabular*}
  \end{minipage}\hfill
  \begin{minipage}[t]{0.48\textwidth}
    \begin{tabular*}{\linewidth}{@{\extracolsep{\fill}}lrr@{}}
      \toprule
      $K$ & DeepSWE & SWE-Verified \\
      \midrule
      8 & 0.915 & 0.988 \\
      16 & 0.952 & 0.988 \\
      32 & 0.951 & 0.906 \\
      \bottomrule
    \end{tabular*}
  \end{minipage}

  \begin{minipage}[t]{0.48\textwidth}
    \caption{\mcqmethod{} scoring-method ablation: Spearman correlation with post-trained SWE-bench Verified on ten base checkpoints.}
    \label{tab:mcq-downstream-app}
  \end{minipage}\hfill
  \begin{minipage}[t]{0.48\textwidth}
    \caption{Prefix-conditioned \passk{} sampling-budget ablation: Spearman correlation with post-trained SWE-bench Verified on ten base checkpoints.}
    \label{tab:passk-k-ablation}
  \end{minipage}
\end{table}

\textbf{A modest sampling budget suffices for prefix-conditioned \passk{}.} \Cref{tab:passk-k-ablation} varies the number of sampled continuations $K$. On held-out DeepSWE prefixes, rank agreement rises from $\rho=0.915$ at $K=8$ to 0.952 at $K=16$ and then plateaus (0.951 at $K=32$). On same-benchmark SWE-bench Verified prefixes it is already 0.988 at $K=8$ and $K=16$ but falls to 0.906 at $K=32$, where pass rates approach saturation (up to 83.7\%) and Nemotron Ultra ties DeepSeek V4 Pro. Larger $K$ therefore does not monotonically improve the screen, and $K=16$ captures nearly all of the attainable agreement on both corpora. Detailed results are in Appendix \ref{app:passk-detail}.

\FloatBarrier
\section{Conclusion}
\label{sec:conclusion}
We introduce a framework for evaluating base checkpoints’ agentic coding potential before committing to expensive post-training. The framework uses successful agent trajectories to provide the context that base checkpoints cannot yet reliably produce themselves and the task’s verifier to identify the decisive step, the first step that resolves the task. Probing checkpoints at the decisive step lets us assess their agentic coding potential without requiring them to operate a tool-use harness across an entire trajectory. Using this framework, we build three complementary screens: \method{}, \mcqmethod{}, and Prefix-Conditioned \passk{}. Across ten base checkpoints, all three produce rankings in close agreement with post-trained SWE-bench Verified performance. The framework thus extends existing and future agentic coding benchmarks into tools for base checkpoint selection before expensive post-training.

\newpage
\bibliographystyle{references}
\bibliography{references}

@misc{openai2026gpt56sol,
  author       = {{OpenAI}},
  title        = {GPT-5.6-Sol},
  howpublished = {\url{https://openai.com/index/gpt-5-6}},
  year         = {2026},
}

@misc{openai2026gpt56terra,
  author       = {{OpenAI}},
  title        = {GPT-5.6-Terra},
  howpublished = {\url{https://openai.com/index/gpt-5-6}},
  year         = {2026},
}

@misc{google2026gemini37flash,
  author       = {{Google}},
  title        = {Gemini 3.7 Flash},
  howpublished = {\url{https://ai.google.dev/gemini-api/docs/models/gemini-3.7-flash}},
  year         = {2026},
}

@misc{google2026gemini36flash,
  author       = {{Google}},
  title        = {Gemini 3.6 Flash},
  howpublished = {\url{https://ai.google.dev/gemini-api/docs/models/gemini-3.6-flash}},
  year         = {2026},
}

@misc{openai2026gpt56luna,
  author       = {{OpenAI}},
  title        = {GPT-5.6-Luna},
  howpublished = {\url{https://openai.com/index/gpt-5-6}},
  year         = {2026},
}

@misc{claude_opus_5_5,
  author       = {{Anthropic}},
  title        = {Claude Opus 5},
  year         = {2026},
  howpublished = {\url{https://www.anthropic.com/news/claude-opus-5}},
}

@misc{claude_opus_4_8,
  author       = {{Anthropic}},
  title        = {Claude Opus 4.8},
  year         = {2026},
  howpublished = {\url{https://www.anthropic.com/news/claude-opus-4-8}},
}

@misc{glm5team2026glm5vibecodingagentic,
      title={GLM-5: from Vibe Coding to Agentic Engineering},
      author={GLM-5-Team and : and Aohan Zeng and Xin Lv and Zhenyu Hou and Zhengxiao Du and Qinkai Zheng and Bin Chen and Da Yin and Chendi Ge and Chenghua Huang and Chengxing Xie and Chenzheng Zhu and Congfeng Yin and Cunxiang Wang and Gengzheng Pan and Hao Zeng and Haoke Zhang and Haoran Wang and Huilong Chen and Jiajie Zhang and Jian Jiao and Jiaqi Guo and Jingsen Wang and Jingzhao Du and Jinzhu Wu and Kedong Wang and Lei Li and Lin Fan and Lucen Zhong and Mingdao Liu and Mingming Zhao and Pengfan Du and Qian Dong and Rui Lu and Shuang-Li and Shulin Cao and Song Liu and Ting Jiang and Xiaodong Chen and Xiaohan Zhang and Xuancheng Huang and Xuezhen Dong and Yabo Xu and Yao Wei and Yifan An and Yilin Niu and Yitong Zhu and Yuanhao Wen and Yukuo Cen and Yushi Bai and Zhongpei Qiao and Zihan Wang and Zikang Wang and Zilin Zhu and Ziqiang Liu and Zixuan Li and Bojie Wang and Bosi Wen and Can Huang and Changpeng Cai and Chao Yu and Chen Li and Chengwei Hu and Chenhui Zhang and Dan Zhang and Daoyan Lin and Dayong Yang and Di Wang and Ding Ai and Erle Zhu and Fangzhou Yi and Feiyu Chen and Guohong Wen and Hailong Sun and Haisha Zhao and Haiyi Hu and Hanchen Zhang and Hanrui Liu and Hanyu Zhang and Hao Peng and Hao Tai and Haobo Zhang and He Liu and Hongwei Wang and Hongxi Yan and Hongyu Ge and Huan Liu and Huanpeng Chu and Jia'ni Zhao and Jiachen Wang and Jiajing Zhao and Jiamin Ren and Jiapeng Wang and Jiaxin Zhang and Jiayi Gui and Jiayue Zhao and Jijie Li and Jing An and Jing Li and Jingwei Yuan and Jinhua Du and Jinxin Liu and Junkai Zhi and Junwen Duan and Kaiyue Zhou and Kangjian Wei and Ke Wang and Keyun Luo and Laiqiang Zhang and Leigang Sha and Liang Xu and Lindong Wu and Lintao Ding and Lu Chen and Minghao Li and Nianyi Lin and Pan Ta and Qiang Zou and Rongjun Song and Ruiqi Yang and Shangqing Tu and Shangtong Yang and Shaoxiang Wu and Shengyan Zhang and Shijie Li and Shuang Li and Shuyi Fan and Wei Qin and Wei Tian and Weining Zhang and Wenbo Yu and Wenjie Liang and Xiang Kuang and Xiangmeng Cheng and Xiangyang Li and Xiaoquan Yan and Xiaowei Hu and Xiaoying Ling and Xing Fan and Xingye Xia and Xinyuan Zhang and Xinze Zhang and Xirui Pan and Xu Zou and Xunkai Zhang and Yadi Liu and Yandong Wu and Yanfu Li and Yidong Wang and Yifan Zhu and Yijun Tan and Yilin Zhou and Yiming Pan and Ying Zhang and Yinpei Su and Yipeng Geng and Yong Yan and Yonglin Tan and Yuean Bi and Yuhan Shen and Yuhao Yang and Yujiang Li and Yunan Liu and Yunqing Wang and Yuntao Li and Yurong Wu and Yutao Zhang and Yuxi Duan and Yuxuan Zhang and Zezhen Liu and Zhengtao Jiang and Zhenhe Yan and Zheyu Zhang and Zhixiang Wei and Zhuo Chen and Zhuoer Feng and Zijun Yao and Ziwei Chai and Ziyuan Wang and Zuzhou Zhang and Bin Xu and Minlie Huang and Hongning Wang and Juanzi Li and Yuxiao Dong and Jie Tang},
      year={2026},
      eprint={2602.15763},
      archivePrefix={arXiv},
      primaryClass={cs.LG},
      url={https://arxiv.org/abs/2602.15763},
}

@article{yang2024sweagent,
  title={Swe-agent: Agent-computer interfaces enable automated software engineering},
  author={Yang, John and Jimenez, Carlos and Wettig, Alexander and Lieret, Kilian and Yao, Shunyu and Narasimhan, Karthik and Press, Ofir},
  journal={Advances in Neural Information Processing Systems},
  volume={37},
  pages={50528--50652},
  year={2024}
}

@misc{claude_sonnet-5,
  author       = {{Anthropic}},
  title        = {Claude Sonnet 5},
  year         = {2026},
  howpublished = {\url{https://www.anthropic.com/news/claude-sonnet-5}},
}

@misc{tencent2026hy3,
  author      = {{Tencent Hy Team}},
  title       = {Hy3},
  institution = {Tencent Hunyuan},
  year        = {2026},
  howpublished = {\url{https://hy.tencent.com/research/hy3}}
}

@article{team2026gemma,
  title={Gemma 4 technical report},
  author={Team, Gemma and Abd, Sherif El and Aggarwal, Vaibhav and Algayres, Robin and Andreev, Alek and Bachem, Olivier and Ballantyne, Ian and Brick, Cormac and C{\u{a}}rbune, Victor and Casbon, Michelle and others},
  journal={arXiv preprint arXiv:2607.02770},
  year={2026}
}

@article{team2026kimi,
  title={Kimi k3: Open frontier intelligence},
  author={Team, Kimi and Bai, Tongtong and Bai, Yifan and Bao, Yiping and Cai, Jianfeng and Cai, Xinyuan and Cao, Peizhou and Cao, Yuxuan and Chai, Ziwei and Charles, Y and others},
  journal={arXiv preprint arXiv:2607.24653},
  year={2026}
}

@article{blakeman2025nemotron,
  title={Nemotron 3 nano: Open, efficient mixture-of-experts hybrid mamba-transformer model for agentic reasoning},
  author={Blakeman, Aaron and Grattafiori, Aaron and Basant, Aarti and Gupta, Abhibha and Khattar, Abhinav and Renduchintala, Adi and Vavre, Aditya and Shukla, Akanksha and Bercovich, Akhiad and Ficek, Aleksander and others},
  journal={arXiv preprint arXiv:2512.20848},
  year={2025}
}

@article{chandiramani2026nemotron,
  title={Nemotron 3 super: Open, efficient mixture-of-experts hybrid mamba-transformer model for agentic reasoning},
  author={Chandiramani, Aakshita and Blakeman, Aaron and Olaoye, Abdullahi and Gupta, Abhibha and Somasamudramath, Abhilash and Khattar, Abhinav and Adesoba, Adeola and Renduchintala, Adi and Asif, Adil and Agrawal, Aditya and others},
  journal={arXiv preprint arXiv:2604.12374},
  year={2026}
}

@article{blakeman2026nemotron,
  title={Nemotron 3 Ultra: Open, Efficient Mixture-of-Experts Hybrid Mamba-Transformer Model for Agentic Reasoning},
  author={Blakeman, Aaron and Thomas, Aaron and Jhunjhunwala, Aastha and Gupta, Abhibha and Khattar, Abhinav and Rajfer, Adam and Renduchintala, Adi and Asif, Adil and Vavre, Aditya and Miranda, Adriana Flores and others},
  journal={arXiv preprint arXiv:2606.15007},
  year={2026}
}

@article{zeng2025glm,
  title={Glm-4.5: Agentic, reasoning, and coding (arc) foundation models},
  author={Zeng, Aohan and Lv, Xin and Zheng, Qinkai and Hou, Zhenyu and Chen, Bin and Xie, Chengxing and Wang, Cunxiang and Yin, Da and Zeng, Hao and Zhang, Jiajie and others},
  journal={arXiv preprint arXiv:2508.06471},
  year={2025}
}

@article{team2025kimi,
  title={Kimi k2: Open agentic intelligence},
  author={Team, Kimi and Bai, Yifan and Bao, Yiping and Charles, Y and Chen, Cheng and Chen, Guanduo and Chen, Haiting and Chen, Huarong and Chen, Jiahao and Chen, Ningxin and others},
  journal={arXiv preprint arXiv:2507.20534},
  year={2025}
}

@misc{qwen3.5,
    title  = {{Qwen3.5}: Towards Native Multimodal Agents},
    author = {{Qwen Team}},
    month  = {February},
    year   = {2026},
    url    = {https://qwen.ai/blog?id=qwen3.5}
}

@misc{deepseekai2026deepseekv4,
      title={DeepSeek-V4: Towards Highly Efficient Million-Token Context Intelligence},
      author={DeepSeek-AI},
      year={2026},
}

@inproceedings{zheng2024large,
  title={Large language models are not robust multiple choice selectors},
  author={Zheng, Chujie and Zhou, Hao and Meng, Fandong and Zhou, Jie and Huang, Minlie},
  booktitle={International Conference on Learning Representations},
  volume={2024},
  pages={19426--19454},
  year={2024}
}

@article{gao2020pile,
  title={The pile: An 800gb dataset of diverse text for language modeling},
  author={Gao, Leo and Biderman, Stella and Black, Sid and Golding, Laurence and Hoppe, Travis and Foster, Charles and Phang, Jason and He, Horace and Thite, Anish and Nabeshima, Noa and others},
  journal={arXiv preprint arXiv:2101.00027},
  year={2020}
}

@article{magnusson2024paloma,
  title={Paloma: A benchmark for evaluating language model fit},
  author={Magnusson, Ian and Bhagia, Akshita and Hofmann, Valentin and Soldaini, Luca and Jha, Ananya H and Tafjord, Oyvind and Schwenk, Dustin and Walsh, Evan and Elazar, Yanai and Lo, Kyle and others},
  journal={Advances in Neural Information Processing Systems},
  volume={37},
  pages={64338--64376},
  year={2024}
}

@inproceedings{chen2025coverage,
  title={The coverage principle: How pre-training enables post-training},
  author={Chen, Fan and Huang, Audrey and Golowich, Noah and Malladi, Sadhika and Block, Adam and Ash, Jordan and Krishnamurthy, Akshay and Foster, Dylan},
  booktitle={International Conference on Learning Representations},
  volume={2026},
  pages={112225--112293},
  year={2026}
}

@article{yue2025does,
  title={Does reinforcement learning really incentivize reasoning capacity in llms beyond the base model?},
  author={Yue, Yang and Chen, Zhiqi and Lu, Rui and Zhao, Andrew and Wang, Zhaokai and Yue, Yang and Song, Shiji and Huang, Gao},
  journal={Advances in Neural Information Processing Systems},
  volume={38},
  pages={57654--57689},
  year={2026}
}

@article{qin2025aptbench,
  title={APTBench: Benchmarking Agentic Potential of Base LLMs During Pre-Training},
  author={Qin, Jiarui and Xi, Yunjia and Huang, Junjie and Rui, Renting and Yin, Di and Liu, Weiwen and Yu, Yong and Zhang, Weinan and Sun, Xing},
  journal={arXiv preprint arXiv:2510.24397},
  year={2025}
}

@article{li2026predicting,
  title={On Predicting the Post-training Potential of Pre-trained LLMs},
  author={Li, Xiaoyuan and Ma, Yubo and Yang, Kexin and Li, Moxin and Bao, Keqin and Wang, Wenie and Feng, Fuli and Liu, Dayiheng},
  journal={arXiv preprint arXiv:2605.11978},
  year={2026}
}

@article{hewitt2024instruction,
  title={Instruction following without instruction tuning},
  author={Hewitt, John and Liu, Nelson F and Liang, Percy and Manning, Christopher D},
  journal={arXiv preprint arXiv:2409.14254},
  year={2024}
}

@article{heineman2025signal,
  title={Signal and noise: A framework for reducing uncertainty in language model evaluation},
  author={Heineman, David and Hofmann, Valentin and Magnusson, Ian and Gu, Yuling and Smith, Noah and Hajishirzi, Hanna and Lo, Kyle and Dodge, Jesse},
  journal={Advances in Neural Information Processing Systems},
  volume={38},
  pages={17073--17114},
  year={2026}
}

@article{chen2021evaluating,
  title={Evaluating large language models trained on code},
  author={Chen, Mark and Tworek, Jerry and Jun, Heewoo and Yuan, Qiming and Pinto, Henrique Ponde De Oliveira and Kaplan, Jared and Edwards, Harri and Burda, Yuri and Joseph, Nicholas and Brockman, Greg and others},
  journal={arXiv preprint arXiv:2107.03374},
  year={2021}
}

@article{austin2021program,
  title={Program synthesis with large language models},
  author={Austin, Jacob and Odena, Augustus and Nye, Maxwell and Bosma, Maarten and Michalewski, Henryk and Dohan, David and Jiang, Ellen and Cai, Carrie and Terry, Michael and Le, Quoc and others},
  journal={arXiv preprint arXiv:2108.07732},
  year={2021}
}

@inproceedings{zhuo2025bigcodebench,
  title={Bigcodebench: Benchmarking code generation with diverse function calls and complex instructions},
  author={Zhuo, Terry Yue and Vu, Minh Chien and Chim, Jenny and Hu, Han and Yu, Wenhao and Widyasari, Ratnadira and Yusuf, Imam Nur Bani and Zhan, Haolan and He, Junda and Paul, Indraneil and others},
  booktitle={International Conference on Learning Representations},
  volume={2025},
  pages={66602--66656},
  year={2025}
}

@inproceedings{jain2025livecodebench,
  title={Livecodebench: Holistic and contamination free evaluation of large language models for code},
  author={Jain, Naman and Gu, Alex and Li, Wen-Ding and Yan, Fanjia and Zhang, Tianjun and Wang, Sida and Solar-Lezama, Armando and Sen, Koushik and Stoica, Ion},
  booktitle={International Conference on Learning Representations},
  volume={2025},
  pages={58791--58831},
  year={2025}
}

@inproceedings{gu2024cruxeval,
  title={{CRUXEval}: A Benchmark for Code Reasoning, Understanding and Execution},
  author={Gu, Alex and Roziere, Baptiste and Leather, Hugh James and Solar-Lezama, Armando and Synnaeve, Gabriel and Wang, Sida},
  booktitle={Proceedings of the 41st International Conference on Machine Learning},
  series={Proceedings of Machine Learning Research},
  volume={235},
  pages={16568--16621},
  year={2024},
}

@inproceedings{liu2024repobench,
  title={{RepoBench}: Benchmarking Repository-Level Code Auto-Completion Systems},
  author={Liu, Tianyang and Xu, Canwen and McAuley, Julian},
  booktitle={International Conference on Learning Representations},
  year={2024},
}

@inproceedings{jimenez2024swebench,
  title={Swe-bench: Can language models resolve real-world github issues?},
  author={Jimenez, Carlos E and Yang, John and Wettig, Alexander and Yao, Shunyu and Pei, Kexin and Press, Ofir and Narasimhan, Karthik},
  booktitle={International Conference on Learning Representations},
  volume={2024},
  pages={54107--54157},
  year={2024}
}

@misc{openai2024verified,
  title={Introducing {SWE-bench Verified}},
  author={{OpenAI}},
  year={2024},
  month={August},
  howpublished={\url{https://openai.com/index/introducing-swe-bench-verified/}},
}

@article{deng2025swebenchpro,
  title={Swe-bench pro: Can ai agents solve long-horizon software engineering tasks?},
  author={Deng, Xiang and Da, Jeff and Pan, Edwin and He, Yannis Yiming and Ide, Charles and Garg, Kanak and Lauffer, Niklas and Park, Andrew and Pasari, Nitin and Rane, Chetan and others},
  journal={arXiv preprint arXiv:2509.16941},
  year={2025}
}

@article{wu2025invisibleleash,
  title={The invisible leash: Why rlvr may or may not escape its origin},
  author={Wu, Fang and Xuan, Weihao and Lu, Ximing and Liu, Mingjie and Dong, Yi and Harchaoui, Zaid and Choi, Yejin},
  journal={arXiv preprint arXiv:2507.14843},
  year={2025}
}

@article{huang2024compression,
  title={Compression represents intelligence linearly},
  author={Huang, Yuzhen and Zhang, Jinghan and Shan, Zifei and He, Junxian},
  journal={arXiv preprint arXiv:2404.09937},
  year={2024}
}

@inproceedings{thrush2024improving,
  title={Improving pretraining data using perplexity correlations},
  author={Thrush, Tristan and Potts, Christopher and Hashimoto, Tatsunori},
  booktitle={International Conference on Learning Representations},
  volume={2025},
  pages={17475--17509},
  year={2025}
}

@misc{holtzman2021surface,
  title={Surface Form Competition: Why the Highest Probability Answer Isn't Always Right},
  author={Holtzman, Ari and West, Peter and Shwartz, Vered and Choi, Yejin and Zettlemoyer, Luke},
  year={2021},
  eprint={2104.08315},
  archivePrefix={arXiv},
  primaryClass={cs.CL},
  url={https://arxiv.org/abs/2104.08315}
}

@article{hendrycks2021apps,
  title={Measuring coding challenge competence with apps},
  author={Hendrycks, Dan and Basart, Steven and Kadavath, Saurav and Mazeika, Mantas and Arora, Akul and Guo, Ethan and Burns, Collin and Puranik, Samir and He, Horace and Song, Dawn and others},
  journal={arXiv preprint arXiv:2105.09938},
  year={2021}
}

@inproceedings{lai2022ds1000,
  title={DS-1000: A natural and reliable benchmark for data science code generation},
  author={Lai, Yuhang and Li, Chengxi and Wang, Yiming and Zhang, Tianyi and Zhong, Ruiqi and Zettlemoyer, Luke and Yih, Wen-tau and Fried, Daniel and Wang, Sida and Yu, Tao},
  booktitle={International Conference on Machine Learning},
  pages={18319--18345},
  year={2023},
  organization={PMLR}
}

@article{liu2023evalplus,
  title={Is your code generated by chatgpt really correct? rigorous evaluation of large language models for code generation},
  author={Liu, Jiawei and Xia, Chunqiu Steven and Wang, Yuyao and Zhang, Lingming},
  journal={Advances in neural information processing systems},
  volume={36},
  pages={21558--21572},
  year={2023}
}

@article{ding2023crosscodeeval,
  title={Crosscodeeval: A diverse and multilingual benchmark for cross-file code completion},
  author={Ding, Yangruibo and Wang, Zijian and Ahmad, Wasi and Ding, Hantian and Tan, Ming and Jain, Nihal and Ramanathan, Murali Krishna and Nallapati, Ramesh and Bhatia, Parminder and Roth, Dan and others},
  journal={Advances in Neural Information Processing Systems},
  volume={36},
  pages={46701--46723},
  year={2023}
}

@article{yang2025swesmith,
  title={Swe-smith: Scaling data for software engineering agents},
  author={Yang, John and Lieret, Kilian and Jimenez, Carlos and Wettig, Alexander and Khandpur, Kabir and Zhang, Yanzhe and Hui, Binyuan and Press, Ofir and Schmidt, Ludwig and Yang, Diyi},
  journal={Advances in Neural Information Processing Systems},
  volume={38},
  year={2026}
}

@misc{lu2026frontiercode,
  title={Introducing {FrontierCode}},
  author={Lu, Eric and Pan, Ben and Birlikci, Deniz and Lee, Sam and Wang, Ray and Choudhury, Rohan and Ma, Fermi and Qin, TC and Baronio, Carlo and Alberti, Silas},
  year={2026},
  howpublished={\url{https://cognition.com/blog/frontier-code}},
}

@article{huang2026deepswe,
  title={DeepSWE: Measuring Frontier Coding Agents on Original, Long-Horizon Engineering Tasks},
  author={Huang, Wenqi and Lee, Charley and Tng, Leonard and Ge, Serena},
  journal={arXiv preprint arXiv:2607.07946},
  year={2026}
}

@inproceedings{merrill2026terminalbench,
  title={{Terminal-Bench}: Benchmarking Agents on Hard, Realistic Tasks in Command Line Interfaces},
  author={Merrill, Mike A. and Shaw, Alexander G. and Carlini, Nicholas and Li, Boxuan and Raj, Harsh and Bercovich, Ivan and Shi, Lin and Shin, Jeong Yeon and Walshe, Thomas and Buchanan, E. Kelly and Shen, Junhong and Ye, Guanghao and Lin, Haowei and Poulos, Jason and Wang, Maoyu and Nezhurina, Marianna and Jitsev, Jenia and Lu, Di and Mastromichalakis, Orfeas Menis and Xu, Zhiwei and others},
  booktitle={International Conference on Learning Representations},
  year={2026},
}

\appendix
\section{Independent Audit of SWE -MCQ Construction}
\label{app:aptbench-audit}
To evaluate whether APTBench's~\citep{qin2025aptbench} SWE MCQs have a uniquely correct designated answer and incorrect distractors, we audited all 1{,}682 questions in four components: Fix Patch, Plan, Test Patch, and Handle Error. We developed a rubric with agent assistance and applied it through automated review, assessing whether the designated answer correctly addresses the stated question and whether another answer choice is also correct. The audit included code execution and inspection of the underlying repository where needed. \cref{tab:aptbench-audit} reports the audit results: 727 questions (43.2\%) were classified as Invalid, meaning that the rendered question does not support the designated answer choice as uniquely correct.

\begin{table}[h]
  \centering
  \caption{Rubric-based assessment of 1{,}682 APTBench SWE MCQs.
An MCQ is classified as Invalid when the question does not
have a uniquely correct designated answer and incorrect distractors.
Percentages are relative to all questions in each row.}
  \label{tab:aptbench-audit}
  \small
  \begin{tabular}{@{}lrrr@{}}
    \toprule
    Component & All questions & Invalid questions & Invalid (\%) \\
    \midrule
    Fix Patch    & 232     & 63  & 27.2\% \\
    Plan         & 243     & 81  & 33.3\% \\
    Test Patch   & 1{,}060 & 450 & 42.5\% \\
    Handle Error & 147     & 133 & 90.5\% \\
    \midrule
    Total        & 1{,}682 & 727 & 43.2\% \\
    \bottomrule
  \end{tabular}
\end{table}

The audit identifies failures of both answer correctness and uniqueness. For example, Handle Error MCQ~\#67 asks how SceneNN data was processed, but its designated correct option recommends reinstalling \texttt{torch\_scatter}, which does not correctly answer the stated question. In Plan MCQ~\#66, answer choices A and C contain identical text, ``I need to run the script again,'' but only A is designated correct. This MCQ therefore accepts answer choice A as correct while rejecting identical answer choice C.

These examples show that selecting the designated answer choice need not mean identifying a uniquely correct answer: scoring can reward an incorrect answer or reject an identical answer that was not designated correct. Such defects weaken the interpretation of MCQ accuracy as the ability to distinguish correct from incorrect answers, even when aggregate scores correlate with post-training performance, as our APTBench evaluation in \cref{app:aptbench-repro} illustrates. These findings motivate \mcqmethod{}'s stricter construction (\cref{sec:mcq-def}): explicit validation of each answer choice. \mcqmethod{} requires the golden action to resolve the task and each distractor to fail the same verifier from the same environment state (\cref{sec:mcq-def}), grounding its answer choices in observed task outcomes.

\section{APTBench Reproduction}
\label{app:aptbench-repro}
We evaluate APTBench on the same ten base checkpoints used to evaluate the three probes, as described in \cref{sec:exp-method}. \Cref{tab:apt-bench-repro} reports Pearson and Spearman correlations between these checkpoints' APTBench scores and the SWE-bench Verified performance of their corresponding post-trained releases.

\begin{table}[h]
\centering
\caption{APTBench reproduction on the same ten base checkpoints evaluated by the three probes (\cref{sec:exp-method}). Pearson and Spearman correlations between APTBench scores and SWE-bench Verified performance of the corresponding post-trained release for each checkpoint.}
\label{tab:apt-bench-repro}
\small
\setlength{\tabcolsep}{4pt}
\begin{tabular}{@{}lrr@{}}
\toprule
& \multicolumn{2}{c}{SWE-bench Verified} \\
\cmidrule(lr){2-3}
APTBench metric & Pearson & Spearman \\
\midrule
APTBench-SWE average & 0.942 & 0.879 \\
APTBench-DR average & 0.962 & 0.903 \\
\addlinespace
APTBench-SWE: EnvSetup Plan ACC & 0.910 & 0.818 \\
APTBench-SWE: EnvSetup Action EM & 0.851 & 0.842 \\
APTBench-SWE: EnvSetup Handle Error ACC & 0.871 & 0.770 \\
APTBench-SWE: IssueFix Locate ACC & 0.457 & 0.377 \\
APTBench-SWE: IssueFix Fix Patch ACC & 0.792 & 0.830 \\
APTBench-SWE: IssueFix Plan ACC & 0.873 & 0.806 \\
APTBench-SWE: IssueFix Action EM & 0.394 & 0.232 \\
APTBench-SWE: IssueFix Test Patch ACC & 0.434 & 0.316 \\
\addlinespace
APTBench-DR: Closed Plan EN ACC & 0.923 & 0.842 \\
APTBench-DR: Closed Plan ZH ACC & 0.903 & 0.855 \\
APTBench-DR: Closed Action EN EM & 0.834 & 0.646 \\
APTBench-DR: Closed Action EN ROUGE-1 & 0.929 & 0.903 \\
APTBench-DR: Closed Action ZH EM & 0.881 & 0.821 \\
APTBench-DR: Closed Action ZH ROUGE-1 & 0.964 & 0.988 \\
APTBench-DR: Open Plan EN ACC & 0.823 & 0.491 \\
APTBench-DR: Open Citation EN ACC & 0.695 & 0.721 \\
APTBench-DR: Open Citation ZH ACC & 0.859 & 0.927 \\
APTBench-DR: Open Action EN ACC & 0.762 & 0.650 \\
APTBench-DR: Open Action ZH ACC & 0.750 & 0.515 \\
\bottomrule
\end{tabular}
\end{table}

\FloatBarrier
\section{Experimental Setup Details}
\label{app:setup-detail}

\begin{table}[!htbp]
  \centering
  \caption{Evaluation cohort and downstream targets: SWE-bench Verified~\citep{jimenez2024swebench,openai2024verified} and Terminal-Bench 2.1~\citep{merrill2026terminalbench}. Each row pairs a base checkpoint with the post-trained release whose reported performance we use as the downstream target; naming it matters because several families shipped more than one. All targets are \passone{} percentages.}
  \label{tab:evaluation-scope}
  \small
  \setlength{\tabcolsep}{4pt}
  \begin{tabular}{@{}llrrr@{}}
    \toprule
    & & \multicolumn{1}{c}{SWE-bench} & \multicolumn{1}{c}{SWE-bench} & \multicolumn{1}{c}{Terminal-Bench} \\
    Base checkpoint & Post-trained model & \multicolumn{1}{c}{Verified} & \multicolumn{1}{c}{Multilingual} & \multicolumn{1}{c}{2.1} \\
    \midrule
    Nemotron-Nano-Base    & Nemotron-Nano & 38.8 & 33.3 & 6.7  \\
    Nemotron-Super-Base   & Nemotron-Super & 60.5 & 45.8 & 38.6 \\
    Nemotron-Ultra-Base   & Nemotron-Ultra & 70.7 & 67.7 & 53.9 \\
    Qwen-3.5-35B-A3B-Base  &  Qwen-3.5-35B-A3B & 69.2 & 60.3 & 48.3 \\
    DeepSeek-V4-Flash-Base& DeepSeek-V4-Flash & 79.0 & 73.3 & 78.7 \\
    Kimi-K2-Base          & Kimi-K2-Thinking & 71.3 & 61.1 & 34.8 \\
    GLM 4.5-Air-Base      & GLM-4.5-Air & 57.6 & 43.2 & 23.6 \\
    Hy3-Preview-Base      & Hy3 & 78.0 & 75.8 & 63.3 \\
    DeepSeek-V4-Pro-Base  & DeepSeek-V4-Pro & 80.6 & 76.2 & 78.7 \\
    Gemma-4-26B           &  Gemma-4-26B-IT & 52.0 & 17.3 & 39.0 \\
    \bottomrule
  \end{tabular}
\end{table}

\paragraph{\method{} corpora and checkpoints.} We evaluate agent trajectories from DeepSWE, SWE-Pro, and SWE-bench Verified on the same ten checkpoints listed in \cref{tab:evaluation-scope}. The primary Decisive-Action BPB results and the all-step versus decisive-step ablation use this matched ten-checkpoint cohort throughout. On DeepSWE, the canonical comparison reported in \cref{fig:teaser,fig:bpb-ablation-summary} increases Spearman correlation with post-trained SWE-bench Verified from 0.915 to 0.964.

\paragraph{\mcqmethod{} cohorts.} \Cref{tab:mcq-base-results-app} uses two four-option question sets, each with chance accuracy 0.250: DeepSWE contributes 756 questions over 392 instances, and SWE-bench Verified 576 questions over 249 instances. Both cover the same ten checkpoints under matched \texttt{none} preambles. We report one accuracy per checkpoint and setting, as recorded in our source measurements; the run-to-run spread behind those figures is not stored alongside them, so we attach no repeat counts or rerun intervals to individual cells and do not read small per-checkpoint differences as measurements of a difference. Detailed construction, coverage, and provenance auditing are reported with the results and in \cref{app:mcq-detail}.

\paragraph{Prefix-conditioned \passk{} cohort.} The sampled probe of \cref{sec:passk-def} replays decisive-step prefixes on 120 DeepSWE instances and on 98 SWE-bench Verified prefixes, over the same ten-checkpoint cohort, at $k \in \{8,16,32\}$ rollouts per instance. Rollouts fork at the decisive step: earlier steps are served verbatim from the captured trajectory, the decisive step is masked, and the candidate checkpoint keeps control for the rest of the episode, with each rollout scored by the task's own benchmark verifier. DeepSWE is cross-domain relative to downstream SWE-bench Verified; the SWE-bench Verified arm uses prefixes and downstream validation on the same benchmark. Because the candidate drives the harness after the fork, these scores are properties of a model--harness system and are not comparable across agent formats.

\paragraph{Statistical protocol.} All reported comparisons use the explicitly stated completed coverage for that experiment: ten models for the primary \method{}, \mcqmethod{}, and prefix-conditioned \passk{} downstream correlations, 99 trajectories for the template intersection, and the per-column available cells for \cref{tab:mcq-base-results-app}. We report Pearson $r$ and Spearman $\rho$ (average ranks for ties) as descriptive point estimates, with no bootstrap intervals available, treating all coefficients as indicative rather than calibrated predictive guarantees.

\section{Evaluation Infrastructure}
\label{app:infra}

\begin{figure}[h]
  \centering
  \includegraphics[width=\textwidth]{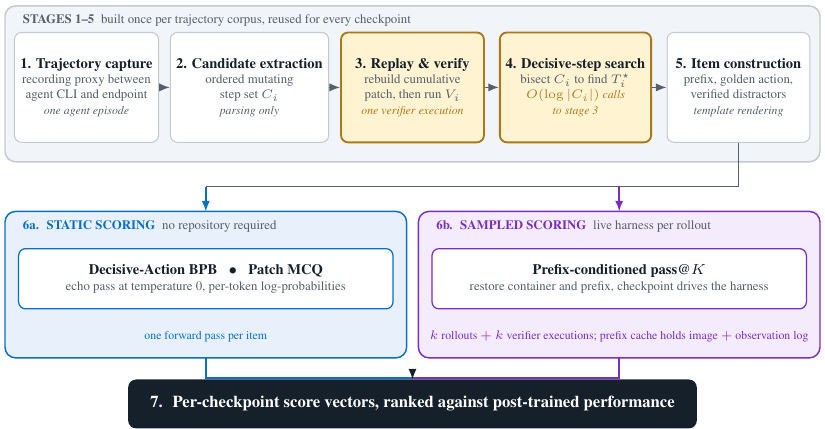}
  \caption{\textbf{The pipeline shared by all three probes.} Stages 1--5 run once per trajectory corpus and produce artifacts reused for every checkpoint; stage 6 is the only per-checkpoint work, and it is where the static and sampled probes diverge by orders of magnitude. Verifier executions, not forward passes, are the scarce resource, which is why stage 4 bisects rather than scans and why the sampled path caches prefixes. }
  \label{fig:infra-pipeline}
\end{figure}

All three probes run on one pipeline and diverge only at the final scoring stage. \Cref{fig:infra-pipeline} shows the stages and what dominates the cost of each. The design constraint throughout is that verifier executions, not forward passes, are the expensive resource: a single candidate step must be replayed into a repository and have the task's own test suite run against it, so the pipeline is organized to bound how many times that happens per trajectory.

\paragraph{Trajectory capture.} A recording proxy sits between the agent CLI and the model endpoint, tees every wire request and response into a local store tagged by (batch, instance), and a post-hoc pass groups those rows into per-step records. Capture is therefore agent-agnostic, which is what lets one pipeline ingest four agent families -- a Responses-API agent, a Messages-API agent, a chat-completions agent, and a host-side agent that executes each shell command through the container -- and recover the family automatically from the captured wire format during replay. It also means the recorded step boundaries are the agent's own, not a reconstruction of them.

\paragraph{Per-instance execution environment.} Every instance runs in its own container built from the benchmark's published task image, with the repository checked out and the interpreter environment the task ships. Before an episode the working tree is reset to the task's base commit with \texttt{git reset --hard} followed by \texttt{git clean -fd}; the \texttt{-x} flag is deliberately omitted, because these images keep compiled extensions and packaging metadata as git-ignored build artifacts, and deleting them leaves a tree that cannot import the package under test. The patch is extracted with \texttt{git add -A} and a binary diff against the base commit, so the same ignore rules keep those artifacts out of the extracted patch. Prompts carry the problem statement only; the benchmark's hint fields are deliberately excluded as a leakage control.

\paragraph{Replay and verification.} Stage 3 is the only component that executes untrusted code. For a candidate step it replays the recorded steps into a fresh container, extracts the cumulative patch, and runs the benchmark's own verifier, recording the replay driver, the patch content or hash, the verifier command and environment, the pass/fail outcome, and an explicit infrastructure-error status. That last field is load-bearing: a verifier that rejects a patch is a normal search observation, whereas a verifier that could not run at all must not be read as rejection, so the two are recorded separately and only the latter invalidates a trial. Replay is strict in one specific way -- a replayed command that does not match the recorded one raises rather than falling through to live execution. Falling through would leave the conversation the candidate reads describing a filesystem that a different command sequence produced, which yields a wrong-but-plausible score rather than an obvious crash. When the candidate takes over, its $k$ rollouts are scored in $k$ separate harness runs, because the harness keys predictions by instance identifier and $k$ patches for one instance would otherwise collide.

\paragraph{Search strategy and its cost.} Stage 4 bisects $C_i$ rather than scanning it, which reduces the verifier executions per trajectory from $|C_i|$ to $O(\log |C_i|)$ and is the single largest infrastructure saving in the pipeline. Bisection assumes monotonicity -- once a cumulative patch passes, later prefixes also pass -- and a linear \texttt{--scan} mode exists for corpora that violate it. Both modes first verify $\max C_i$ as a gate, so a trajectory whose final step does not resolve is discarded after one verifier execution instead of an entire search. All decisive-step metadata reported in this paper was produced by bisect.

\paragraph{Throughput limits and the failure mode they create.} Both scoring paths are endpoint-bound rather than client-bound: raising client concurrency on a saturated endpoint changed static-scoring throughput not at all, while two hosts differed by more than an order of magnitude on identical work. Shared endpoints also impose a concurrency ceiling, and exceeding it fails in a way that is not self-announcing. Past the ceiling the retry budget exhausts, a rate-limited forward returns nothing, and the rollout falls back to a prefix-only patch that is indistinguishable from a legitimate failed attempt -- a degraded window reads as a low score rather than as an error. We observed one candidate score zero under saturation and 0.347 on a clean re-run, with the size of the correction tracking each arm's measured rate-limit fraction. Two safeguards follow: runs are capped below the measured ceiling, and empty predictions written during a degraded window are purged rather than resumed over, since resumption logic keyed on file existence would otherwise treat them as complete forever. Long prefixes are dropped when they exceed an endpoint's context limit, and because the longest prefixes are also the most-explored instances, those drops are a selection effect and comparisons are made on the completed intersection (\cref{app:setup-detail}).

\paragraph{Sampling configuration.} For base checkpoints driving the harness, sampling temperature is 0.6. 

\section{Bounded Code Benchmarks as Baselines}
\label{app:bounded-baselines}

\Cref{sec:intro} argues that bounded code benchmarks -- those a base checkpoint can complete without sustained interaction with an evolving repository -- do not measure whether it can become a capable coding agent. \Cref{tab:bounded-baselines-app} quantifies that claim on the ten-checkpoint cohort of \cref{tab:evaluation-scope} by correlating each benchmark's base-model score with the same two post-trained downstream targets used throughout. BigCodeBench~\citep{zhuo2025bigcodebench} is named in \cref{sec:intro} as a member of this family but is not among our measurements, so it does not appear here; the two LiveCodeBench variants and the two CRUXEval directions are reported separately because they score different abilities.

\begin{table}[h]
  \centering
  \small
  \caption{Bounded code benchmarks as baselines for post-trained agentic performance, with our three probes repeated for reference. Correlations are against each post-trained downstream target on the cohort of \cref{tab:evaluation-scope}; the probe rows use the DeepSWE-derived probes of \cref{fig:deepswe-probes-swe-verified,fig:deepswe-probes-terminal}, with $-\operatorname{BPB}$ for \method{}. Two baselines cover only ten checkpoints, so their coefficients are not matched pairs with the rest. At $n \le 10$ with no interval estimates these are descriptive associations, and single-row differences should not be over-read.}
  \label{tab:bounded-baselines-app}
  \begin{tabular}{@{}lcrrrr@{}}
    \toprule
    & & \multicolumn{2}{c}{vs.\ SWE-bench Verified} & \multicolumn{2}{c}{vs.\ Terminal-Bench 2.1} \\
    \cmidrule(lr){3-4}\cmidrule(lr){5-6}
    Evaluation & $n$ & Pearson & Spearman & Pearson & Spearman \\
    \midrule
    MBPP & 10 & 0.151 & 0.091 & 0.037 & 0.036 \\
    HumanEval & 10 & $-0.369$ & $-0.394$ & $-0.650$ & $-0.571$ \\
    LiveCodeBench & 10 & 0.523 & 0.455 & 0.429 & 0.383 \\
    CRUXEval-I & 10 & 0.473 & 0.394 & 0.287 & 0.201 \\
    CRUXEval-O & 10 & 0.814 & 0.758 & 0.618 & 0.523 \\
    RepoBench XFirst & 10 & 0.690 & \textbf{0.830} & 0.557 & 0.505 \\
    \midrule
    \method{}, decisive step & 10 & 0.965 & 0.964 & 0.863 & 0.833 \\
    \mcqmethod{}, think 1000 & 10 & 0.915 & 0.867 & 0.871 & 0.796 \\
    Prefix-conditioned \passk{}, $K=32$ & 10 & 0.917 & 0.951 & 0.947 & 0.930 \\
    \bottomrule
  \end{tabular}
\end{table}

\paragraph{The family spans from negatively ranked to moderate, and none matches our method.} RepoBench XFirst is the strongest bounded baseline by rank agreement ($\rho=0.830$), which is the figure \cref{sec:intro} and \cref{sec:conclusion} quote as the comparison point for \method{}. CRUXEval-O follows and is genuinely informative ($r=0.814$, $\rho=0.758$): predicting the output of a program exercises reasoning about execution, which is closer to what resolving a repository issue demands than synthesizing a standalone function is. The synthesis benchmarks supply almost no signal on this cohort -- MBPP is flat ($\rho=0.091$) and HumanEval is \emph{negatively} ranked ($\rho=-0.394$) -- and the pattern is consistent with range compression rather than noise alone: across these ten checkpoints MBPP spans only 67.2--86.0 and RepoBench XFirst 73.2--82.8, while the downstream target spans 38.8--80.6. A benchmark whose scores are bunched near the top cannot order checkpoints that differ by more than forty points downstream.

\paragraph{What this does and does not establish.} Two readings are unsupported. First, these coefficients do not show that bounded benchmarks measure nothing: CRUXEval-O and RepoBench XFirst both carry real signal, and a practitioner with no trajectory corpus would be better off consulting them than nothing. Second, the comparison is not controlled. The baselines are single published-style scores, our probes are computed on trajectories we selected, and every downstream score comes from a different lab's post-training recipe and harness. What the table supports is narrower and sufficient for \cref{sec:intro}: within one cohort, agreement with post-trained agentic performance varies enormously across bounded benchmarks, the best of them still trails all three trajectory-derived probes, and nothing in the family distinguishes itself in a way that would let a practitioner pick one in advance.

\section{End-to-End Evaluation of Base Checkpoints}
\label{app:e2e}
\FloatBarrier

If base checkpoints could simply be run end-to-end on an agentic coding benchmark, the probes in \cref{sec:method} would be unnecessary. To assess the need for the proposed probes, we evaluate base models directly in an end-to-end setting. We run six base checkpoints on SWE-bench Verified, driving the  harness from the task statement (with no trajectory context), and results are list in \cref{tab:e2e}.

\begin{table}[h]
  \centering
  \small
  \caption{End-to-end evaluation of base checkpoints on SWE-bench Verified, alongside each family's post-trained score on the same benchmark. All values are \passone{} percentages.}
  \label{tab:e2e}
  \begin{tabular}{@{}lrr@{}}
    \toprule
    & \multicolumn{1}{c}{End-to-end} & \multicolumn{1}{c}{Post-trained} \\
    Base checkpoint & \multicolumn{1}{c}{\passone{}} & \multicolumn{1}{c}{SWE-Verified} \\
    \midrule
    DeepSeek-V4-Pro-Base   & 0.0  & \textbf{80.6} \\
    DeepSeek-V4-Flash-Base & 0.0  & 79.0 \\
    Nemotron-Ultra-Base    & 0.0  & 70.7 \\
    Qwen-3.5-35B-A3B-Base  & \textbf{20.2} & 69.2 \\
    Nemotron-Super-Base    & 0.0  & 60.5 \\
    Nemotron-Nano-Base     & 0.0  & 38.8 \\
    \bottomrule
  \end{tabular}
\end{table}

\noindent\textbf{End-to-end evaluation cannot rank base checkpoints, and the one signal it gives points the wrong way.} Five of the six checkpoints resolve no task at all, so the benchmark ties them at the floor and cannot order them. Only Qwen-3.5-35B-A3B-Base resolves any, at 20.2\%, which ranks it first end to end even though its post-trained family is fourth of the six, behind both DeepSeek V4 models and Nemotron Ultra. End-to-end \passone{} therefore ties five of the six checkpoints at the floor, and its one non-zero score places a mid-ranked family first: it is not merely uninformative but, where it discriminates at all, misleading. The exception of Qwen-3.5-35B-A3B-Base more plausibly reflects its familiarity with agent-formatted interaction than coding ability.

\section{Correlations with Terminal-Bench 2.1}
\label{app:terminal-bench}

\begin{figure}[!htbp]
  \centering
  \includegraphics[width=0.9\textwidth]{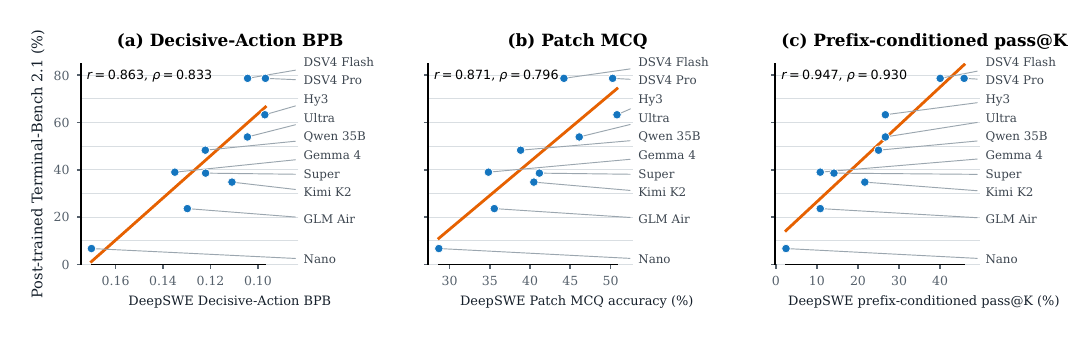}
  \caption{\textbf{DeepSWE prefix-conditioned \passk{} at $K=32$ has the strongest descriptive agreement with post-trained Terminal-Bench 2.1 performance.} Panels use the same ten checkpoints and the same probe settings as \cref{fig:deepswe-probes-swe-verified}; annotations report Pearson $r$ and Spearman $\rho$. The BPB axis is reversed so rightward consistently means better.}
  \label{fig:deepswe-probes-terminal}
\end{figure}

\Cref{fig:deepswe-probes-terminal} tests whether the DeepSWE-derived probes of \cref{sec:exp-swemult-tb-corr} also predict post-trained performance on Terminal-Bench 2.1, which scores command-line workflows rather than repository issues. On Terminal-Bench 2.1, the prefix-conditioned \passk{} transfers best. Agreement is lower for the two static probes, at $\rho=0.833$ for \method{} and 0.796 for \mcqmethod{}, while prefix-conditioned \passk{} holds at 0.930. 

\noindent\textbf{Where the Terminal-Bench gap comes from.} Two checkpoints account for most of the disagreement between the benchmarks. Kimi K2 ranks fourth on SWE-bench Verified but eighth on Terminal-Bench, and Gemma 4 26B ranks ninth and sixth. The probes side with the SWE view of both, placing Kimi K2 fifth or sixth and Gemma 4 26B eighth or ninth, which is what costs them agreement with Terminal-Bench. The shortfall is thus a property of how the two benchmarks rank these models rather than a failure of the probes on the SWE ranking they were built from. Capturing Terminal-Bench-specific standing would likely require trajectories from command-line tasks, and aggregating probes across trajectory sources from several task families is a natural way to do so.

\section{\method{} Method Detail}
\label{app:method-detail}

\subsection{Universal Chat Template and Formatting Masks}
\label{app:chat-template}

\paragraph{Universal rendering.}
We serialize every trajectory with the same model-independent plain-text template rather than using each checkpoint's native chat template. Messages remain in their original chronological order, and the renderer uses the following visible labels  as shown in \cref{tab:universal-chat-template}.

\begin{table}[h]
  \centering
  \caption{Universal serialization of trajectory components. Angle-bracketed text denotes the original message content.}
  \label{tab:universal-chat-template}
  \small
  \begin{tabular}{@{}p{0.30\linewidth}p{0.61\linewidth}@{}}
    \toprule
    Source component & Universal representation \\
    \midrule
    Tool definitions & \texttt{Tool definitions: <JSON>} \\
    System message & \texttt{System: <content>} \\
    User message & \texttt{User: <content>} \\
    Assistant text & \texttt{Assistant: <content>} \\
    Assistant tool call & \texttt{Assistant (tool call): <JSON>} \\
    Tool response & \texttt{Tool Result: <content>} \\
    \bottomrule
  \end{tabular}
\end{table}
\FloatBarrier

An assistant message may contain both explanatory text and one or more tool calls; in that case, both the \texttt{Assistant:} and \texttt{Assistant (tool call):} records are emitted. The template contains no model-specific control tokens such as ChatML markers or model-specific step delimiters: all labels above are ordinary text.

\paragraph{Canonical tool-call serialization.}
Tool definitions and tool calls are serialized with \texttt{json.dumps(\allowbreak value, ensure\_ascii=False)}. This preserves source field insertion order, uses the standard spacing after commas and colons, and emits Unicode directly. If \texttt{function.arguments} is already a JSON string, it remains a string inside the outer tool-call object rather than being parsed into a nested object; reparsing would change the prediction target and make it artificially easier. Random tool-call IDs are removed before rendering because they carry no semantic information and are not predictably determined by the trajectory. All other tool-call fields are preserved.

\paragraph{Prefix and target construction.}
For an assistant step at position $i$, the prefix is the universal rendering of the tool definitions and messages strictly before $i$, while the target is the same rendering applied to message $i$ alone. Tool definitions therefore appear in the prefix but are not repeated in the target. Using the same serializer on both sides prevents a representation mismatch at the prefix--target boundary.

\paragraph{Formatting mask.}
The complete rendered prefix and decisive response are visible to the checkpoint as conditioning context, but the primary \method{} result accumulates loss only over assistant-generated semantic content: natural-language assistant text, tool function names, and tool arguments. It excludes fixed role labels; tool-call wrapper keys such as \texttt{type}, \texttt{function}, \texttt{name}, and \texttt{arguments}; JSON braces, brackets, quotes, commas, and colons belonging to that wrapper; and trailing record separators. Our tool-formatting-included ablation adds the serialized tool-call wrapper to the scored region, but still excludes the fixed role labels. The rendered input is otherwise identical, so the two settings differ only in which target bytes contribute to the BPB numerator and denominator.

\section{Patch MCQ Detail}
\label{app:mcq-detail}

\begin{table}[h]
  \centering
  \caption{Source-current four-option \mcqmethod{} accuracy with cyclic rotation and arithmetic aggregation; chance is 0.250. Both question sets use matched \texttt{none} preambles. Each cell is the one accuracy our source measurements record for that checkpoint and setting; run-to-run spread is not stored with these values (\cref{app:setup-detail}), so column-to-column gaps of a few thousandths should not be read as differences. Bold marks the largest value per column.}
  \label{tab:mcq-base-results-app}
  \footnotesize
  \begin{tabular}{@{}lrrrr@{}}
    \toprule
    & \multicolumn{2}{c}{DeepSWE} & \multicolumn{2}{c}{SWE-Verified} \\
    \cmidrule(lr){2-3}\cmidrule(lr){4-5}
    Base checkpoint & No think & Think & No think & Think \\
    \midrule
    Hy3-Preview-Base & \textbf{0.4456} & \textbf{0.5080} & 0.552 & 0.583 \\
    DeepSeek-V4-Pro-Base & 0.4206 & 0.5026 & 0.550 & \textbf{0.591} \\
    Nemotron-Ultra-Base & 0.4235 & 0.4611 & 0.538 & 0.558 \\
    DeepSeek-V4-Flash-Base & 0.4009 & 0.4421 & 0.536 & 0.553 \\
    Kimi-K2-Base & 0.4322 & 0.4047 & \textbf{0.597} & 0.523 \\
    Qwen-3.5-35B-A3B-Base & 0.3953 & 0.3883 & 0.521 & 0.524 \\
    Nemotron-Super-Base & 0.3472 & 0.4116 & 0.500 & 0.498 \\
    GLM 4.5-Air-Base & 0.3513 & 0.3556 & 0.425 & 0.432 \\
    Gemma-4-26B & 0.3265 & 0.3483 & 0.443 & 0.440 \\
    Nemotron-Nano-Base & 0.3172 & 0.2867 &  0.306 & 0.342 \\
    \midrule
    Mean & 0.386 & 0.411 & 0.497 & 0.504 \\
    \bottomrule
  \end{tabular}
\end{table}

\begin{table}[h]
  \centering
  \small
  \caption{\mcqmethod{} read ablation on both question sets; chance is 0.250. Argmin-BPB scores each option independently and picks the lowest byte-normalized negative log-likelihood; rotated log-probability is the answer-letter read, presenting all four options jointly under four cyclic rotations, and is reported here at think-1000 to match \cref{tab:mcq-base-results-app}. Gap is argmin-BPB minus rotated log-probability, so negative values favor the joint read. Rows are ordered by DeepSWE rotated log-probability.}
  \label{tab:mcq-read-ablation}
  \begin{tabular}{@{}lrrrrrr@{}}
    \toprule
    & \multicolumn{3}{c}{DeepSWE} & \multicolumn{3}{c}{SWE-Verified} \\
    \cmidrule(lr){2-4}\cmidrule(lr){5-7}
    Base checkpoint & Argmin & Rotated & Gap & Argmin & Rotated & Gap \\
    \midrule
    Hy3-Preview-Base & 0.266 & 0.508 & $-0.242$ & 0.359 & 0.583 & $-0.224$ \\
    DeepSeek-V4-Pro-Base & 0.283 & 0.503 & $-0.220$ & 0.405 & 0.591 & $-0.186$ \\
    Nemotron-Ultra-Base & 0.267 & 0.461 & $-0.194$ & 0.366 & 0.558 & $-0.192$ \\
    DeepSeek-V4-Flash-Base & 0.266 & 0.442 & $-0.176$ & 0.380 & 0.553 & $-0.173$ \\
    Kimi-K2-Base & 0.279 & 0.405 & $-0.126$ & 0.387 & 0.523 & $-0.136$ \\
    Qwen-3.5-35B-A3B-Base & 0.271 & 0.388 & $-0.117$ & 0.325 & 0.524 & $-0.199$ \\
    Nemotron-Super-Base & 0.269 & 0.412 & $-0.143$ & 0.366 & 0.498 & $-0.132$ \\
    GLM 4.5-Air-Base & 0.265 & 0.356 & $-0.091$ & 0.384 & 0.432 & $-0.048$ \\
    Gemma-4-26B & 0.262 & 0.348 & $-0.086$ & 0.354 & 0.440 & $-0.086$ \\
    Nemotron-Nano-Base & 0.262 & 0.287 & $-0.025$ & 0.347 & 0.342 & $+0.005$ \\
    \midrule
    Mean & 0.269 & 0.411 & $-0.142$ & 0.367 & 0.504 & $-0.137$ \\
    \bottomrule
  \end{tabular}
\end{table}

\begin{table}[h]
  \centering
  \small
  \caption{Single-rotation versus full cyclic rotation on the DeepSWE question set, think-1000 arm, over the same ten checkpoints as \cref{tab:mcq-base-results-app}; chance is 0.250. $k{=}1$ takes the answer from the first rotation alone and $k{=}4$ is the shipped aggregate, both recomputed offline from the same finished runs, so no checkpoint was re-scored; each value is the mean of three rounds. The $k{=}4$ column is the DeepSWE think-1000 column of \cref{tab:mcq-base-results-app}. Rotation helps on 9 of 10 checkpoints, but not uniformly: the gain spans $-0.005$ to $+0.081$, wider than the gap between adjacent checkpoints in the $k{=}4$ ranking, and it reorders the top two. Bold marks the best value per column.}
  \label{tab:mcq-rotation-ablation}
  \begin{tabular}{@{}lrrr@{}}
    \toprule
    Base checkpoint & $k{=}1$ & $k{=}4$ & Gain \\
    \midrule
    DeepSeek-V4-Pro-Base & \textbf{0.4555} & 0.5026 & $+0.047$ \\
    Hy3-Preview-Base & 0.4271 & \textbf{0.5080} & \textbf{$+0.081$} \\
    Nemotron-Ultra-Base & 0.4195 & 0.4611 & $+0.042$ \\
    DeepSeek-V4-Flash-Base & 0.3881 & 0.4421 & $+0.054$ \\
    Kimi-K2-Base & 0.3741 & 0.4047 & $+0.031$ \\
    Nemotron-Super-Base & 0.3727 & 0.4116 & $+0.039$ \\
    Qwen-3.5-35B-A3B-Base & 0.3454 & 0.3883 & $+0.043$ \\
    GLM 4.5-Air-Base & 0.3333 & 0.3556 & $+0.022$ \\
    Gemma-4-26B & 0.3126 & 0.3483 & $+0.036$ \\
    Nemotron-Nano-Base & 0.2913 & 0.2865 & $-0.005$ \\
    \midrule
    Mean & 0.3720 & 0.4109 & $+0.039$ \\
    \bottomrule
  \end{tabular}
\end{table}

\begin{table}[h]
  \centering
  \small
  \caption{Downstream agreement of every \mcqmethod{} scoring method on the same ten checkpoints, with both Pearson and Spearman correlations. Rows pair a question set with a scoring method; the SWE-Verified question set draws its options from the same benchmark as the first target, so those rows are same-benchmark evidence against it while all rows are cross-domain with respect to Terminal-Bench 2.1. At $n=10$ with no interval estimates we do not read small differences as separating rows.}
  \label{tab:mcq-downstream-full-app}
  \begin{tabular}{@{}llrrrr@{}}
    \toprule
    & & \multicolumn{2}{c}{vs.\ SWE-bench Verified} & \multicolumn{2}{c}{vs.\ Terminal-Bench 2.1} \\
    \cmidrule(lr){3-4}\cmidrule(lr){5-6}
    Question set & Scoring method & Pearson & Spearman & Pearson & Spearman \\
    \midrule
    DeepSWE & Argmin-BPB & 0.601 & 0.665 & 0.420 & 0.382 \\
    DeepSWE & Rotated, no think & 0.900 & 0.806 & 0.686 & 0.547 \\
    DeepSWE & Rotated, think 1000 & 0.915 & 0.867 & 0.871 & 0.796 \\
    \midrule
    SWE-Verified & Argmin-BPB & 0.429 & 0.578 & 0.359 & 0.213 \\
    SWE-Verified & Rotated, no think & 0.909 & 0.842 & 0.725 & 0.578 \\
    SWE-Verified & Rotated, think 1000 & 0.971 & 0.903 & 0.894 & 0.900 \\
    \bottomrule
  \end{tabular}
\end{table}

\begin{figure}[h]
  \centering
  \includegraphics[width=0.9\textwidth]{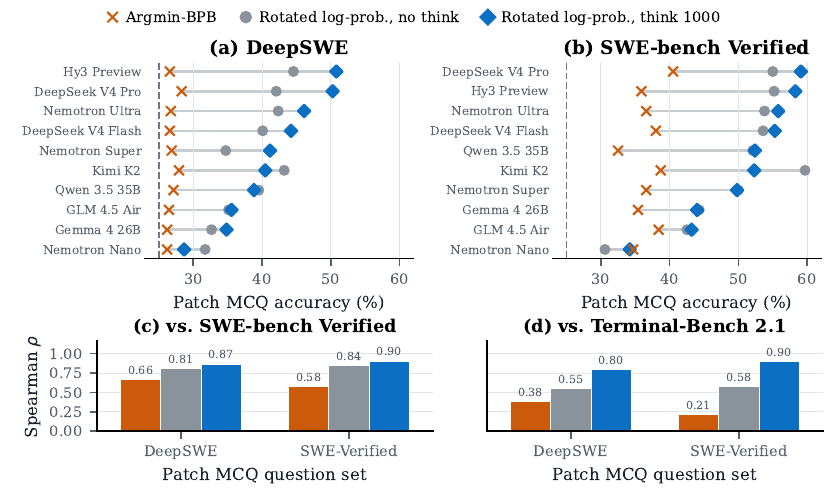}
  \caption{\textbf{\mcqmethod{} accuracy and downstream agreement, by scoring rule.} Panels (a) and (b) rank the same ten checkpoints on each question set; panels (c) and (d) plot the rank agreement tabulated in \cref{tab:mcq-downstream-full-app} against both downstream targets. The joint answer-letter read separates the cohort more sharply than scoring each option independently, and the gap between the two reads is larger than the gap between thinking protocols.}
  \label{fig:mcq-ranking-app}
\end{figure}

\paragraph{Question-set construction detail.} \Cref{tab:mcq-base-results-app}'s DeepSWE question set is built by a max-flow selection over three source sets capped at two questions per instance. It contains 756 questions over 392 instances; the four gold-option length ranks and four answer slots each occur exactly 189 times, so the pick-longest baseline is exactly 0.250. The SWE-Verified question set merges the earlier pair pool with two additional single-action rollout waves, deduplicates by source, instance, target step, and trial, and retains 576 length-rebalanced questions over 249 instances.

\paragraph{Presentation and answer extraction.} Each question $q_i$ is presented under four cyclic option reorderings (rotations), indexed $r \in \{0,1,2,3\}$ with $r=0$ the original ordering. Each reordering is an independent request whose MCQ context $u_i^{(r)}$ contains the trajectory prefix and all four options, so the options are always compared within one shared context as in \cref{eq:mcq-letter-likelihood}. Let $\ell_r(o) \in \mathcal{L}_4$ denote the answer letter under which option $o$ is displayed in reordering $r$; because the reorderings are cyclic, every option is displayed under every letter exactly once across the four requests. From each request we read the next-token probability of each answer letter, restricted to the top-20 candidate tokens the endpoint returns, and renormalize over the answer letters,
\begin{equation}
\tilde p_\theta\big(\ell \mid u_i^{(r)}\big) := \frac{p_\theta\big(\ell \mid u_i^{(r)}\big)}{\sum_{\ell' \in \mathcal{L}_4} p_\theta\big(\ell' \mid u_i^{(r)}\big)}, \qquad \ell \in \mathcal{L}_4,
\end{equation}
assigning zero probability to any letter absent from the returned candidates. Because a given letter names a different option in each reordering, these probabilities are aggregated per option rather than per letter: each option's aggregate is the mean probability of the letter it was displayed under, taken over the four reorderings,
\begin{equation}
\bar p_i(o) := \frac{1}{4} \sum_{r=0}^{3} \tilde p_\theta\big(\ell_r(o) \mid u_i^{(r)}\big),
\label{eq:mcq-aggregate}
\end{equation}
and the prediction is the option with the largest aggregate, reported by its letter in the original ordering so that it is directly comparable with the golden option $\bar\ell_i$ in the accuracy of \cref{sec:mcq-def}:
\begin{equation}
\widehat{\ell}_i := \ell_0\Big(\arg\max_{o}\, \bar p_i(o)\Big),
\end{equation}
with exact ties broken by position in the original ordering. With a single reordering this reduces to \cref{eq:mcq-letter-likelihood}. Averaging probabilities, rather than counting per-reordering votes, keeps a near-tie between two options from being rounded away before the four reads are combined, and it cancels the position prior because that prior falls equally on every option over a full cycle. With reasoning enabled, $u_i^{(r)}$ additionally contains the checkpoint's own reasoning of up to 1{,}000 tokens, and the letter probabilities are read at the answer position that follows it; reconstructing the answer from the distribution before the reasoning describes a different quantity and yields near-chance accuracy.

\paragraph{Only complete reorderings are scored.} A reordering fails to read if none of the answer letters appears among the returned candidates, and a question counts toward accuracy only if all four of its reorderings read successfully; a question with any failed reordering is left unscored, since \cref{eq:mcq-aggregate} is defined only over a complete cycle. This is a correctness requirement rather than a cleanliness preference. Rotation removes the position prior by making it common to every option so that it cancels in the comparison, and that cancellation is exactly what an incomplete set of rotations destroys: a partially rotated item still carries the prior and therefore measures a different quantity than a fully rotated one. Admitting partial rows changes per-checkpoint results by amounts comparable to the effects we report, so every arm in this section is filtered to complete rotations.

\paragraph{Rotation count.} \Cref{tab:mcq-rotation-ablation} isolates how much the joint read owes to rotation itself. Taking the answer from the first rotation alone, rather than aggregating all four, costs 0.039 accuracy averaged over the cohort, and the cost is uneven: it ranges from $-0.005$ on Nemotron Nano, the one checkpoint rotation does not help, to $+0.081$ on Hy3 Preview. That spread is wider than the gap between adjacent checkpoints in the four-rotation ranking, and it is enough to reorder the top two: DeepSeek V4 Pro leads on a single read (0.4555 versus 0.4271) but trails on the aggregate (0.5026 versus 0.5080). Part of a checkpoint's standing under \mcqmethod{} therefore reflects how much the debiasing protocol recovers for it, not its per-read judgement alone, so the read protocol belongs with the score whenever \mcqmethod{} numbers are quoted.

\paragraph{Argmin-BPB scoring ablation.} \Cref{tab:mcq-read-ablation} compares the two scoring methods on both question sets, with the rotated-log-probability columns taken at think-1000 so they match \cref{tab:mcq-base-results-app} exactly. The joint read wins on all ten checkpoints on DeepSWE and on nine of ten on SWE-Verified, where Nano is the lone exception by 0.005. The two reads also rank the cohort differently, at Spearman $\rho=0.543$ on DeepSWE and $0.322$ on SWE-Verified, and argmin-BPB's agreement with post-trained SWE-bench Verified \passone{} is correspondingly weaker ($\rho=0.665$ and $0.578$, versus $0.867$ and $0.903$ for the joint read; \cref{tab:mcq-downstream-full-app} reports every scoring method against both downstream targets, and \cref{fig:mcq-ranking-app} plots the same comparison), so the choice of scoring method changes probing conclusions and not just absolute levels.

\paragraph{Reasoning effects require matched protocols.} Both question sets now provide format-matched \texttt{none}-preamble comparisons. On DeepSWE, the ten-checkpoint mean rises from 0.386 to 0.411 ($+0.025$): seven checkpoints improve, with changes ranging from $+0.082$ for DeepSeek V4 Pro to $-0.030$ for Nano. On SWE-Verified the mean rises only from 0.497 to 0.504 ($+0.008$), also with 7/10 improving, and Kimi drops by 0.074 -- a single checkpoint accounting for most of the difference between the two means. Without run-to-run spread for these values we do not treat small per-checkpoint differences as measured effects, and no-think and think-1000 remain distinct protocols rather than one uniformly dominating the other.

\FloatBarrier
\section{Prefix-Conditioned \texorpdfstring{\passk{}}{pass@k} Detail}
\label{app:passk-detail}

\paragraph{Fork point and masking convention.} The decisive-step index is inclusive: masking at $T_i^\star$ means steps $1,\ldots,T_i^\star-1$ are replayed verbatim from the recorded trajectory and the candidate must itself produce the action at $T_i^\star$. Replaying one step further and generating nothing reproduces the recorded resolving patch exactly, which we use as a control that the replay path is faithful before any candidate is scored.

\paragraph{Forward budget.} Unlike a single-action evaluation, the candidate has to finish the task after the fork, so its phase is multi-turn and needs a stopping rule; every reported run caps it at three generated steps. When that budget is exhausted the proxy returns a synthetic response carrying no action, so the agent exits cleanly and the patch is still extracted, rather than the episode running to timeout and yielding nothing.

\begin{figure}[htbp]
  \centering
  \includegraphics[width=\textwidth]{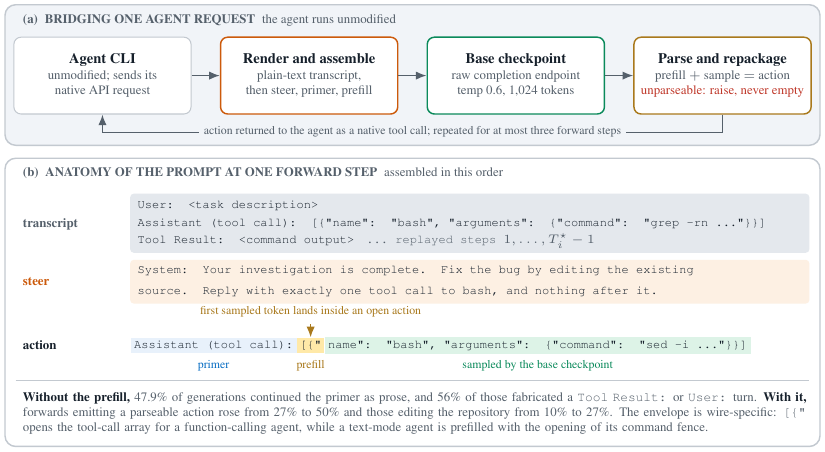}
  \caption{\textbf{Driving an agent harness from a base checkpoint.} (a) The bridge sits between an unmodified agent CLI and the base checkpoint: it renders each agent request into a plain-text transcript, samples a continuation from the checkpoint's raw completion endpoint, and returns the parsed action to the agent as a native tool call. (b) The prompt at one forward step, assembled as transcript, steer, primer, and prefill. The prefill pre-opens the action envelope so that the checkpoint's first sampled token already lands inside an open action; only the green span is written by the base checkpoint.}
  \label{fig:base-model-bridge}
\end{figure}

\paragraph{Driving the harness from a base checkpoint.} A post-trained candidate is relayed: the agent's own request is forwarded to it unchanged. A base checkpoint cannot be driven that way, because it has no chat template and cannot reliably emit a structured action, so we bridge it onto the interface instead. The bridge intercepts each request the agent CLI makes, renders it into a role-labeled plain-text transcript using the same renderer as \method{} (\cref{app:chat-template}), samples a continuation from the checkpoint's raw completion endpoint, and repackages the result as the response the agent expects, so the agent itself runs unmodified. \Cref{fig:base-model-bridge} shows this round trip and the prompt it produces. The prompt is assembled in a fixed order,
\[
\text{transcript} \;\Vert\; \text{steer} \;\Vert\; \text{primer} \;\Vert\; \text{prefill},
\]
and the order matters as much as the contents. The \emph{steer} is a final \texttt{System:} line stating that the investigation is complete and asking for exactly one action that edits the existing source; given only a neutral primer, a base checkpoint tends to keep exploring or to write a reproduction script instead of a fix. The \emph{primer} is the turn header \texttt{Assistant (tool call):}, which places the next token at the start of an action. Placing the steer after the primer instead of before it cut the share of forwards emitting a parseable action to 12\%.

\paragraph{Prefilling the action envelope.} The primer alone is not enough, and prefilling is what makes a base checkpoint usable in the harness at all. With only the primer, 47.9\% of generations on one checkpoint continued it as prose rather than acting, and 56\% of those fabricated a \texttt{Tool Result:} or \texttt{User:} turn: the checkpoint kept writing the transcript instead of acting in it. Adding the steer did not fix this, since 79\% of the generations that still failed to parse simply echoed the steer back as another \texttt{System:} block. The fix is to append the opening of the action envelope after the primer, so that the checkpoint's first sampled token already lands inside an open action and a prose continuation becomes ungrammatical; the fragment is prepended back onto the sample before parsing, so the parser sees a complete action. On eight DeepSWE prefixes with 48 generations per arm, prefilling raised the share of forwards emitting a parseable action from 27\% to 50\%, raised the share that edit the repository from 10\% to 27\%, and cut steer echoes from 19\% to 8\%. The envelope is wire-specific and must open exactly what the agent parses: for the function-calling agent on DeepSWE it is \texttt{[\{"}, the start of the tool-call JSON array, and for the text-mode agent on SWE-bench Verified it is the opening of the fenced command block that agent reads. Stop sequences follow the wire in the same way: they are kept in text mode, where they are load-bearing, and disabled for the function-calling agent.

\paragraph{Reading the action back.} The completion is parsed for the checkpoint's own action with no instruct model in the loop: nothing repairs malformed output, so what is scored is exactly what the base checkpoint wrote. A turn that cannot be parsed raises an error rather than returning an empty action, because the agent would read an empty action as the episode being finished, and the rollout would then end looking like a legitimate failed attempt rather than a formatting failure. Each forward is sampled at temperature 0.6 (\cref{app:infra}) with a 1{,}024-token cap: at 512 tokens, 95\% of generations hit the cap and were truncated mid-command, and raising it to 1{,}024 lifted the parse rate from 68\% to 80\%. Because prefilling forces the first token to open an action, it leaves the checkpoint no room to deliberate, so we also tested a two-step variant that samples a short plan before the action. It lowered the share of forwards emitting a parseable action from 56\% to 38\%, since the spliced plan gave the checkpoint more transcript to continue instead of acting; every reported run therefore samples each action directly, with no plan step, within the three-step forward budget.

\paragraph{Prefix caching.} The $k$ rollouts at one decisive step replay identical prefix steps, which on a 352-rollout measurement accounted for 48\% of rollout wall time. Caching that work requires two artifacts, because a prefix is two kinds of state: a committed container image carries the filesystem the candidate edits, and an observation log carries the conversation it reads. Restoring only the image saves nothing, since a host-side agent rebuilds its message history by re-executing commands; restoring only the log is worse, because the candidate then edits a pristine tree while reading a history that says otherwise. On ten prefixes the cache reproduced byte-identical patches and cut prefix wall time by 96\%, roughly 40\% per arm over the full pool, at about 64\,GB of images per 120 prefixes. Committing a container preserves the filesystem but drops running processes, so prefixes whose recorded commands leave a background service behind are detected by probing both containers and dropped from the cache, with the uncached path used as the fallback.

\paragraph{Scoring the rollouts.} The $k$ rollouts for one instance are scored in $k$ separate harness runs, because the benchmark harness keys predictions by instance identifier and $k$ patches for one instance would otherwise collide in a single predictions file. Generation is endpoint-bound while the harness is CPU- and disk-bound, so the two are pipelined: one rollout index is generated while the previous index is scored in the background. We validated the pipelined path as byte-identical to the serial one before using it for the reported runs.

\begin{table}[htbp]
  \centering
  \caption{Prefix-conditioned \passk{} on DeepSWE and SWE-bench Verified trajectory prefixes, in percent. DeepSWE uses 120 held-out instances; SWE-bench Verified uses a separately eligibility-filtered cohort of 98 prefixes. Both cover the same ten checkpoints, but the two halves aggregate over different cohort sizes, so their means are not matched pairs. Rollouts fork at the decisive step $T_i^\star$: steps before are served verbatim, $T_i^\star$ is masked, and the checkpoint keeps control for the rest of the episode. An instance counts as resolved if at least one of $k$ continuations passes the verifier. Row order follows \cref{tab:mcq-base-results-app}; bold marks the largest value per column.}
  \label{tab:passk-results-app}
  \footnotesize
  \begin{tabular}{@{}lrrrrrr@{}}
    \toprule
    & \multicolumn{3}{c}{DeepSWE} & \multicolumn{3}{c}{SWE-bench Verified} \\
    \cmidrule(lr){2-4}\cmidrule(lr){5-7}
    Base checkpoint & $k=8$ & $k=16$ & $k=32$ & $k=8$ & $k=16$ & $k=32$ \\
    \midrule
    \texttt{hy3-base} & 16.67 & 20.83 & 26.67 & 61.22 & 70.41 & 76.53 \\
    \texttt{deepseek-v4-pro-base} & \textbf{37.50} & \textbf{41.67} & \textbf{45.83} & \textbf{68.37} & \textbf{77.55} & \textbf{83.67} \\
    \texttt{ultra-base} & 12.50 & 17.50 & 26.67 & 58.16 & 68.37 & \textbf{83.67} \\
    \texttt{dsflash-base} & 28.33 & 33.33 & 40.00 & 63.27 & 72.45 & 79.59 \\
    \texttt{kimi-k2-base} & 14.17 & 16.67 & 21.67 & 55.10 & 62.24 & 68.37 \\
    \texttt{qwen-base} & 17.50 & 20.00 & 25.00 & 52.04 & 57.17 & 63.27 \\
    \texttt{super-base} & 5.83 & 10.83 & 14.17 & 41.84 & 50.00 & 58.16 \\
    \texttt{glm-air-base} & 8.33 & 9.17 & 10.83 & 34.69 & 42.86 & 51.02 \\
    \texttt{gemma-base} & 4.17 & 6.67 & 10.83 & 31.63 & 39.80 & 47.96 \\
    \texttt{nano-base} & 1.67 & 2.50 & 2.50 & 18.37 & 26.53 & 35.71 \\
    \midrule
    Mean & 14.67 & 17.92 & 22.42 & 48.47 & 56.74 & 64.80 \\
    \bottomrule
  \end{tabular}
\end{table}

\begin{table}[htbp]
  \centering
  \caption{Downstream validation of prefix-conditioned \passk{} against both targets of \cref{tab:evaluation-scope}, on the same ten-checkpoint cohort used throughout the primary comparisons; higher is better agreement. SWE-bench Verified prefix rows are same-benchmark evidence against the SWE-bench Verified target, not cross-domain validation; both prefix corpora are cross-domain with respect to Terminal-Bench 2.1. The last two rows repeat the canonical all-step/decisive-step comparison from \cref{fig:teaser,fig:bpb-ablation-summary}, correlated against $-\operatorname{BPB}$ on the same ten checkpoints. No value is marked best: at $n=10$ with no bootstrap intervals, we do not read these differences as separating the rows.}
  \label{tab:passk-validation-app}
  \footnotesize
  \setlength{\tabcolsep}{3pt}
  \begin{tabular}{@{}l@{\hspace{4pt}}c@{\hspace{6pt}}rr@{\hspace{8pt}}rr@{}}
    \toprule
    & & \multicolumn{2}{c}{vs.\ SWE-bench Verified} & \multicolumn{2}{c}{vs.\ Terminal-Bench 2.1} \\
    \cmidrule(lr){3-4}\cmidrule(lr){5-6}
    Probe & $n$ & Pearson & Spearman & Pearson & Spearman \\
    \midrule
    DeepSWE prefix \passk{}, $k=8$ & 10 & 0.840 & 0.915 & 0.870 & 0.802 \\
    DeepSWE prefix \passk{}, $k=16$ & 10 & 0.872 & 0.952 & 0.912 & 0.900 \\
    DeepSWE prefix \passk{}, $k=32$ & 10 & 0.917 & 0.951 & 0.947 & 0.930 \\
    SWE-bench Verified prefix \passk{}, $k=8$ & 10 & 0.990 & 0.988 & 0.902 & 0.875 \\
    SWE-bench Verified prefix \passk{}, $k=16$ & 10 & 0.984 & 0.988 & 0.911 & 0.875 \\
    SWE-bench Verified prefix \passk{}, $k=32$ & 10 & 0.949 & 0.906 & 0.886 & 0.820 \\
    \midrule
    \method{}, all steps (reference) & 10 & 0.960 & 0.915 & 0.861 & 0.809 \\
    \method{}, decisive step only (reference) & 10 & 0.965 & 0.964 & 0.863 & 0.833 \\
    \bottomrule
  \end{tabular}
\end{table}

\begin{figure}[!t]
  \centering
  \includegraphics[width=0.9\textwidth]{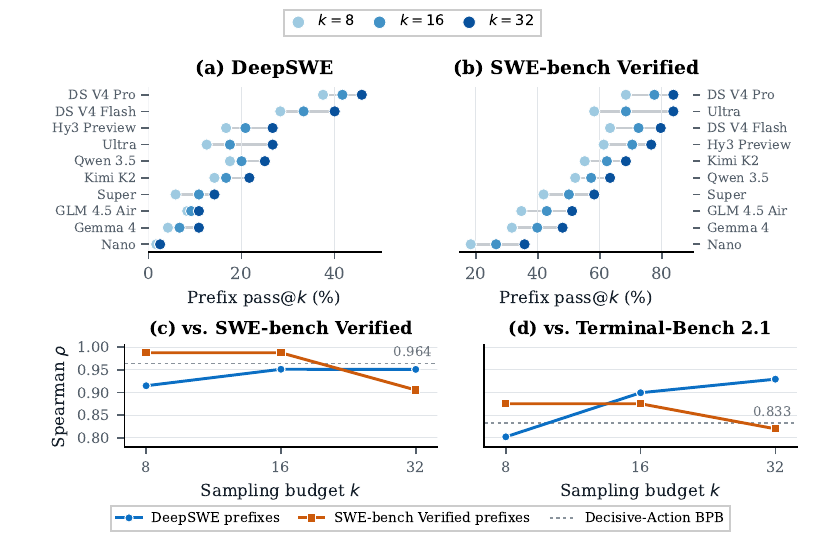}
  \caption{\textbf{Prefix-conditioned \passk{} rankings and downstream agreement.} Panels (a) and (b) rank the same ten checkpoints under DeepSWE and SWE-bench Verified trajectory prefixes at $k \in \{8,16,32\}$, tabulated in \cref{tab:passk-results-app}. The two panels are sorted independently and use different $x$ ranges, so marker positions are not comparable across them. Panels (c) and (d) plot the rank agreement of those scores with each downstream target from \cref{tab:passk-validation-app}; the dashed line marks \method{} against the same target for reference.}
  \label{fig:passk-combined}
\end{figure}

\paragraph{Consistency on one previously mis-ranked pair, but not independent evidence.} On held-out DeepSWE prefixes, prefix-conditioned \passk{} ranks DeepSeek V4 Flash above Nemotron Ultra 550B at every $k$ (40.00\% versus 26.67\% at $k=32$), matching the correction made by Decisive-Action BPB. This is a consistency check rather than an independent correction, because the protocol already forks at the decisive step. The second BPB case, Qwen 3.5 35B versus Kimi K2, remains a disagreement: at $k=32$ the sampled probe places Qwen above Kimi (25.00\% versus 21.67\%) even though Kimi scores 2.1 points higher downstream. The SWE-bench Verified arm uses prefixes from the same benchmark as the downstream target and is not held out in that sense.

\paragraph{Where the probes disagree.} \Cref{fig:passk-combined} shows both arms against both targets. At $k=32$, the held-out DeepSWE arm disagrees with downstream SWE-bench Verified \passone{} on two of 45 checkpoint pairs, both involving Kimi K2, which the sampled probe places below Nemotron Ultra 550B (21.67\% versus 26.67\%) and below Qwen 3.5 35B (21.67\% versus 25.00\%) despite Kimi scoring higher downstream, by 0.6 and 2.1 SWE-bench Verified points respectively. The SWE-bench Verified prefix arm disagrees on three pairs, all involving Nemotron Ultra, which it ranks above DeepSeek V4 Flash (83.67\% versus 79.59\%), Kimi K2 (83.67\% versus 68.37\%), and Hy3 Preview (83.67\% versus 76.53\%) despite Nemotron scoring lower downstream in each case. Kimi also leads the SWE-bench Verified no-think \mcqmethod{} setting in \cref{tab:mcq-base-results-app}, so the cheap and sampled probes disagree about this checkpoint in opposite directions across narrow downstream gaps. This is precisely the case we reserve for direct investigation rather than for arbitration by either proxy.

\section{Additional SFT Correlation Results}
\label{app:sft-corr}

\Cref{tab:probes-sft-corr} shows correlations between our probes and the post-SFT SWE-bench Verified scores of Nemotron 3 Nano, Nemotron 3 Super and Qwen 3.5 35B A3B, using different SWE trajectories as sources. The BPB rows use the same canonical per-checkpoint measurements as the main ten-checkpoint analysis. At $n=3$, one pairwise inversion changes Spearman correlation by 0.5, so these coefficients are descriptive; extending the controlled SFT experiment to more base models is left to future work.

\begin{table}[h]
  \centering
  \small
  \begin{tabular}{@{}lrrrrrr@{}}
    \toprule
    & \multicolumn{2}{c}{vs.\ Post-SFT SWE-Verified}
    & \multicolumn{2}{c}{vs.\ Public SWE-Verified}
    & \multicolumn{2}{c}{vs.\ Public SWE-Verified} \\
    & \multicolumn{2}{c}{($n=3$)}
    & \multicolumn{2}{c}{($n=3$)}
    & \multicolumn{2}{c}{($n=10$)} \\
    \cmidrule(lr){2-3}\cmidrule(lr){4-5}\cmidrule(lr){6-7}
    Evaluation & Pearson & Spearman & Pearson & Spearman & Pearson & Spearman \\
    \midrule
    \multicolumn{7}{@{}l}{\textbf{DeepSWE}} \\
    \quad\method{} & 0.97 & 0.50 & 0.96 & 0.50 & 0.97 & 0.96 \\
    \quad\mcqmethod{} & 0.91 & 0.50 & 0.90 & 0.50 & 0.87 & 0.80 \\
    \quad Prefix pass@32 & 0.97 & 1.00 & 0.98 & 1.00 & 0.95 & 0.93 \\
    \addlinespace
    \multicolumn{7}{@{}l}{\textbf{SWE-Pro}} \\
    \quad\method{} & 0.99 & 1.00 & 0.99 & 1.00 & 0.94 & 0.96 \\
    \quad\mcqmethod{} & 1.00 & 1.00 & 1.00 & 1.00 & 0.66 & 0.61 \\
    \quad Prefix pass@32 & 1.00 & 1.00 & 1.00 & 1.00 & 0.85 & 0.83 \\
    \addlinespace
    \multicolumn{7}{@{}l}{\textbf{SWE-bench Verified}} \\
    \quad\method{} & 0.99 & 1.00 & 0.99 & 1.00 & 0.95 & 0.92 \\
    \quad\mcqmethod{} & 0.99 & 1.00 & 0.99 & 1.00 & 0.89 & 0.90 \\
    \quad Prefix pass@32 & 1.00 & 1.00 & 0.99 & 1.00 & 0.89 & 0.82 \\
    \bottomrule
  \end{tabular}
  \caption{Comparison of probe correlations across source tasks and post-SFT SWE-bench Verified results for Nemotron 3 Nano, Nemotron 3 Super, and Qwen 3.5 35B A3B. We show correlations against public SWE-bench Verified scores for the same 3 models, as well as all 10 models in our panel.}
  \label{tab:probes-sft-corr}
\end{table}

\end{document}